\documentclass[11pt, letterpaper]{template}
\usepackage[authoryear, round]{natbib}
\usepackage{hyperref}[citecolor=magenta]

\hypersetup{
    colorlinks=true,
    citecolor=blue,
    linkcolor=blue,
    urlcolor=blue
}

\definecolor{darkblue}{rgb}{0, 0, 0.5}
\usepackage{url}            % simple URL typesetting
\usepackage{booktabs}       % professional-quality tables
\usepackage{amsfonts}       % blackboard math symbols
\usepackage{nicefrac}       % compact symbols for 1/2, etc.
\usepackage{microtype}      % microtypography
\usepackage{xcolor}         % colors

\usepackage{lineno}

\usepackage{enumitem}       % [leftmargin=20pt]
\usepackage{subfig}         % subfloat
\usepackage{multirow}       % 
\usepackage{xspace}         % xspace
\usepackage{wrapfig}        % wrapfigure
\usepackage{makecell}       % divide row in table
\usepackage{colortbl}
\usepackage{bbm}            % indicator function (\mathbbm)
\usepackage{adjustbox}

\usepackage[most,skins,theorems]{tcolorbox}
\tcbset{
  aibox/.style={
    width=\linewidth,
    top=8pt,
    bottom=4pt,
    colback=blue!6!white,
    colframe=black,
    colbacktitle=black,
    enhanced,
    center,
    attach boxed title to top left={yshift=-0.1in,xshift=0.15in},
    boxed title style={boxrule=0pt,colframe=white,},
  }
}
\newtcolorbox{AIbox}[2][]{aibox,title=#2,#1}

\newcommand{\PRM}{PRM\xspace}
\newcommand{\PRMs}{PRMs\xspace}

\newcommand{\ORMs}{ORMs\xspace}

\newcommand{\tms}{$\times$}
\newcommand{\mean}[1]{#1}

\newcommand\blfootnote[1]{%
  \begingroup
  \renewcommand\thefootnote{}\footnote{#1}%
  \addtocounter{footnote}{-1}%
  \endgroup
}

\title{Can 1B LLM Surpass 405B LLM? Rethinking Compute-Optimal Test-Time Scaling}

\author[1,2,$*$]{Runze Liu}
\author[1,3]{Junqi Gao}
\author[4]{Jian Zhao}
\author[2]{Kaiyan Zhang}
\author[2]{Xiu Li}
\author[1,$\dag$]{Biqing Qi}
\author[1]{Wanli Ouyang}
\author[1,2,$\dag$]{Bowen Zhou}

\affil[1]{Shanghai AI Laboratory}
\affil[2]{Tsinghua University}
\affil[3]{Harbin Institute of Technology}
\affil[4]{BUPT}
\begin{abstract}
Test-Time Scaling (TTS) is an important method for improving the performance of Large Language Models (LLMs) by using additional computation during the inference phase. However, current studies do not systematically analyze how policy models, Process Reward Models (PRMs), and problem difficulty influence TTS. This lack of analysis limits the understanding and practical use of TTS methods. In this paper, we focus on two core questions: \textbf{(1)} What is the optimal approach to scale test-time computation across different policy models, PRMs, and problem difficulty levels? \textbf{(2)} To what extent can extended computation improve the performance of LLMs on complex tasks, and can smaller language models outperform larger ones through this approach? Through comprehensive experiments on MATH-500 and challenging AIME24 tasks, we have the following observations: \textbf{(1)} The compute-optimal TTS strategy is highly dependent on the choice of policy model, PRM, and problem difficulty. \textbf{(2)} With our compute-optimal TTS strategy, extremely small policy models can outperform larger models. For example, a \textbf{1B} LLM can exceed a \textbf{405B} LLM on MATH-500. Moreover, on both MATH-500 and AIME24, a \textbf{0.5B} LLM outperforms \textbf{GPT-4o}, a \textbf{3B} LLM surpasses a \textbf{405B} LLM, and a \textbf{7B} LLM beats \textbf{o1} and \textbf{DeepSeek-R1}, while with \textbf{higher} inference efficiency. These findings show the significance of adapting TTS strategies to the specific characteristics of each task and model and indicate that TTS is a promising approach for enhancing the reasoning abilities of LLMs. Our website is available at \url{https://ryanliu112.github.io/compute-optimal-tts}.
\end{abstract}

\begin{document}

\blfootnote{$^*$ Work done during an internship at Shanghai AI Laboratory}
\blfootnote{$^\dag$ Corresponding authors: Biqing Qi (qibiqing@pjlab.org.cn), Bowen Zhou (zhoubowen@tsinghua.edu.cn)}

\maketitle

% \begin{flushright}
% \textit{xxxxxxxxxxxxxxxxxxxxxxx \\ xxxxxxxxxxxxxxxxxxxxx \\ -Proverb}
% \end{flushright}

\vspace{-0.3em}
\begin{figure*}[!h]
\centering
\vspace{-0.3em}
\includegraphics[width=0.85\linewidth]{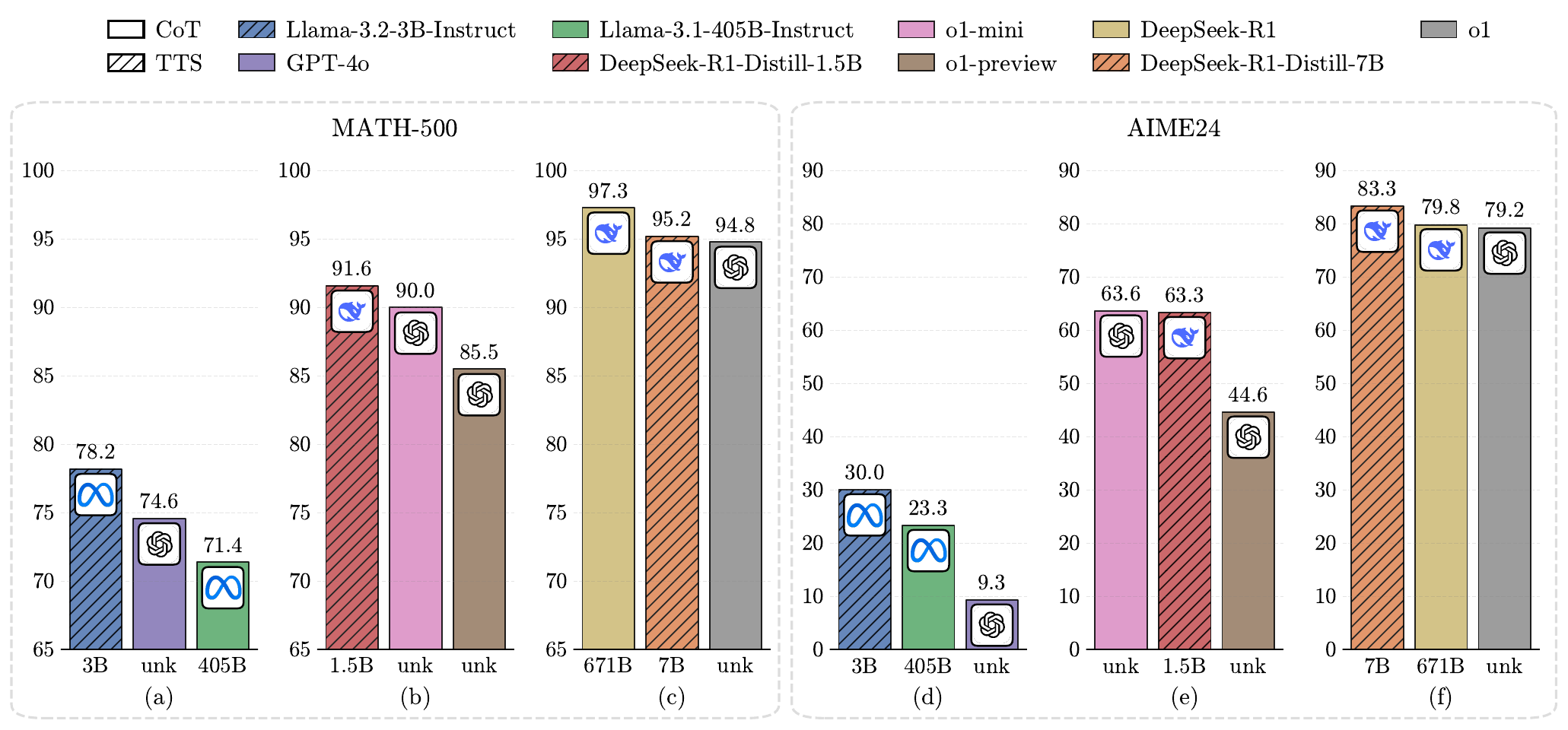}
\vspace{-0.3em}
\caption{Comparison between the performance of smaller LLMs compute-optimal TTS and that of larger LLMs CoT on MATH-500 and AIME24. \textbf{(a) \& (d)} Llama-3.2-3B-Instruct surpasses Llama-3.1-405B-Instruct and GPT-4o on MATH-500 and AIME24; \textbf{(b) \& (e)} DeepSeek-R1-Distill-1.5B outperforms o1-preview on MATH-500 and AIME24, and surpasses o1-mini on MATH-500; \textbf{(c) \& (f)} DeepSeek-R1-Distill-7B beats o1 on MATH-500 and AIME24, and exceeds DeepSeek-R1 on AIME24.
}
\label{fig:compute-optimal}
\end{figure*}

\section{Introduction}
Large Language Models~(LLMs) have shown significant improvements across a variety of domains~\citep{GPT-4, GPT-4o, Claude, o1, DeepSeek-R1}. Recently, OpenAI o1~\citep{o1} has demonstrated that Test-Time Scaling~(TTS) can enhance the reasoning capabilities of LLMs by allocating additional computation at inference time, making it an effective approach for improving LLM performance~\citep{QwQ, Kimi-k1.5, DeepSeek-R1}.

TTS approaches can be divided into two main categories: (1) Internal TTS, which trains the LLMs to ``think'' slowly with long Chain-of-Thought~(CoT)~\citep{o1, DeepSeek-R1}, and (2) External TTS, which improves the reasoning performance via sampling or search-based methods with fixed LLMs~\citep{wu2024inference, snell2024scaling}.
The key challenge of external TTS is how to scale compute optimally, that is, allocating the optimal computation for each problem~\citep{snell2024scaling}. Current TTS methods guide the generation process and select the final answer using Process Reward Models~(PRMs), which effectively scale test-time compute~\citep{wu2024inference, snell2024scaling, huggingface2024scaling}. These TTS methods involve several important factors, such as policy models\footnote{Following~\citet{snell2024scaling}, we use ``policy models'' to refer to LLMs that generate solutions, and ``verifiers'' for \PRMs.}, PRMs, and problem difficulty levels. However, there is limited systematic analysis of how policy models, PRMs, and problem difficulty influence these TTS strategies. This limitation prevents the community from fully understanding the effectiveness of this method and developing insights for compute-optimal TTS strategies.

To address these issues, this paper aims to investigate the influence of policy models, PRMs, and problem difficulty on TTS through comprehensive experimental analysis. Furthermore, we explore the concrete characteristics and performance boundaries of TTS methods. Specifically, we conduct extensive experiments on MATH-500~\citep{PRM800K} and the challenging AIME24~\citep{AIME24} tasks using a range of PRMs (spanning from 1.5B to 72B across different model series) across multiple policy models (ranging from 0.5B to 72B across two model families). Our results show that the compute-optimal TTS strategy heavily depends on the specific policy model, PRM, and problem difficulty level. Even smaller models (e.g., a \textbf{1B} model) can outperform larger models (e.g., a \textbf{405B} model) and even state-of-the-art reasoning models, such as \textbf{o1} or \textbf{DeepSeek-R1}, in challenging reasoning tasks by applying compute-optimal TTS.

The contributions of this work can be summarized as follows:
\begin{enumerate}
    \item We conduct a comprehensive evaluation of different TTS methods using various up-to-date policy models, multiple PRMs, diverse scaling methods, and more challenging tasks.
    \item Our analysis highlights the necessity of considering the influence of rewards in the TTS process and introduces reward-aware compute-optimal TTS. We also demonstrate that the compute-optimal scaling strategy varies with different policy models, PRMs, and problem difficulty levels.
    \item The empirical results demonstrate the significant potential of smaller language models to outperform larger models through TTS. Using the reward-aware Compute-optimal TTS strategy, we show that a \textbf{3B} LLM can outperform a \textbf{405B} LLM, and a \textbf{7B} LLM can surpass \textbf{o1} and \textbf{DeepSeek-R1} on MATH-500 and AIME24 tasks.
\end{enumerate}

\section{Setup \& Preliminaries}

\subsection{Problem Formulation}

We formulate the reasoning problem as a Markov Decision Process (MDP)~\citep{sutton2018reinforcement}, defined by the tuple $(\mathcal{S}, \mathcal{A}, \mathcal{P}, \mathcal{R}, \gamma)$, where $\mathcal{S}$ is the state space, $\mathcal{A}$ is the action space, $\mathcal{P}: \mathcal{S} \times \mathcal{A} \rightarrow \mathcal{S}$ is the transition function, $\mathcal{R}: \mathcal{S} \times \mathcal{A} \rightarrow \mathbb{R}$ is the reward function, and $\gamma \in [0, 1]$ is the discount factor. Given a prompt $x \sim \mathcal{X}$, the policy with parameters $\theta$ generates the initial action $a_1 \sim \pi_\theta(\cdot \mid s_1)$, where $s_1 = x$ is the initial state. The policy receives a reward $\mathcal{R}(s_1, a_1)$, and the state transitions to $s_2 = [s_1, a_1]$, where $[\cdot, \cdot]$ denotes the concatenation of two strings. This process continues until the episode terminates, either by reaching the maximum number of steps or by generating an \texttt{<EOS>} token. A trajectory of length $H$ is represented as $\tau = \{a_1, a_2, \cdots, a_H\}$. The process can be summarized as follows:
\begin{equation}
\begin{array}{ll}
    \text{Initial State:}    & s_1 = x \sim \mathcal{X} \\
    \text{Action:}           & a_t \sim \pi_\theta(\cdot \mid s_t) \\
    \text{State Transition:} & s_{t+1} = \mathcal{P}(\cdot \mid s_t, a_t) = [s_t, a_t] \\
    \text{Reward:}           & r_t = \mathcal{R}(s_t, a_t) \\
\end{array}
\end{equation}

% The objective is to 

% \begin{equation}
% \max ?
% \end{equation}

\subsection{Test-Time Scaling Method}

\begin{figure}[t]
\centering
\includegraphics[width=0.85\linewidth]{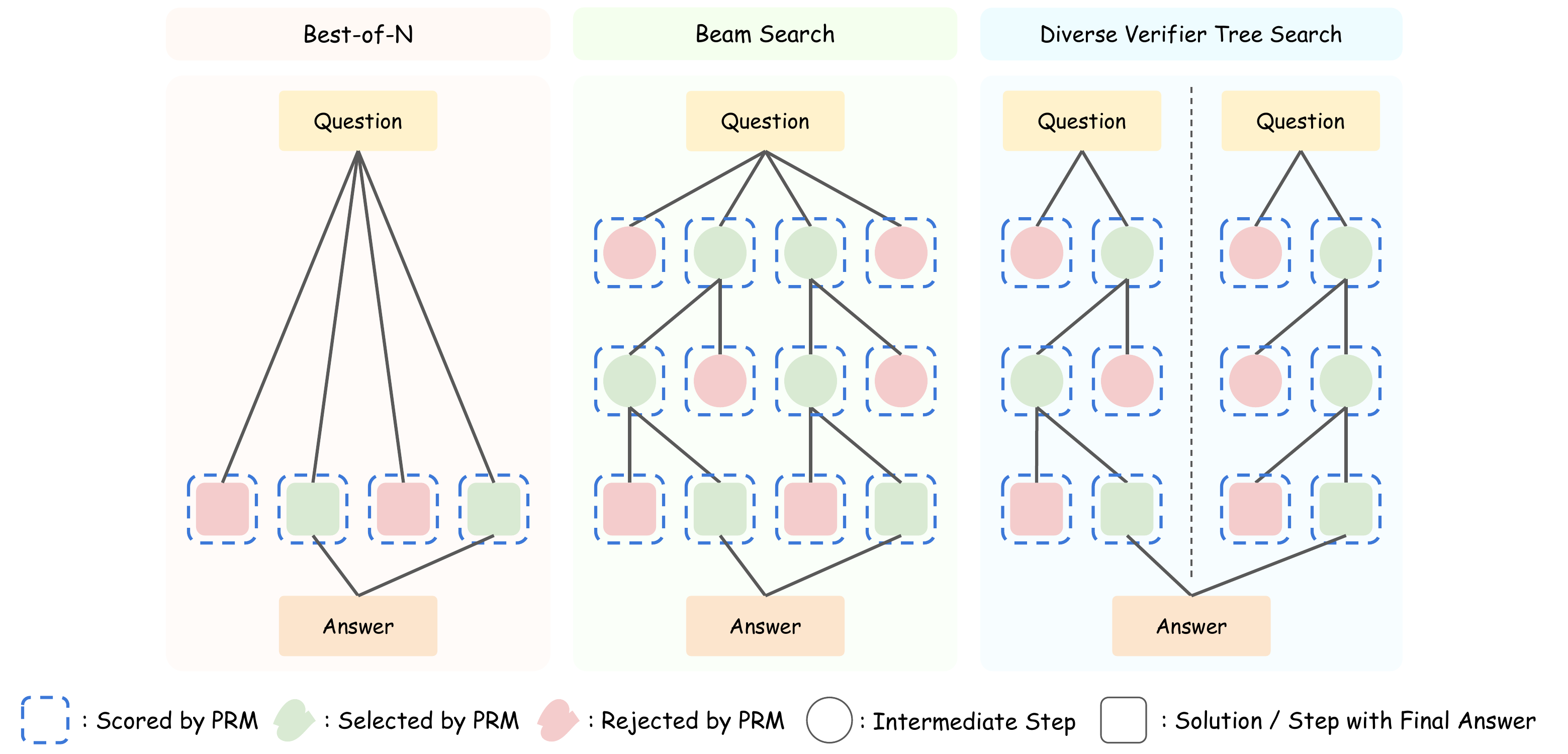}
\caption{Comparison of different external TTS methods.}
\label{fig:tts_method}
\end{figure}

We consider three TTS methods: Best-of-N~(BoN)~\citep{brown2024large}, beam search~\citep{snell2024scaling}, and Diverse Verifier Tree Search~(DVTS)~\citep{huggingface2024scaling}. As pointed out by~\citet{snell2024scaling}, lookahead search is inefficient due to multi-step sampling, so we do not evaluate it or other methods involving lookahead operations, such as Monte Carlo Tree Search~(MCTS). The TTS methods are shown in Figure~\ref{fig:tts_method}.

\paragraph{Best-of-N.}
In the BoN approach, the policy model generates $N$ responses, after which scoring and voting methods are applied to select the final answer.

\paragraph{Beam Search.}
Given beam width $N$ and beam size $M$, the policy model first generates $N$ steps. The verifier selects the top $\frac{N}{M}$ steps for subsequent search. In the next step, the policy model samples $M$ steps for each selected previous step. This process repeats until the maximum depth is reached or an \texttt{<EOS>} token is generated.
% The detailed beam search algorithm can be found in Appendix~\ref{alg:bs}.

\paragraph{Diverse Verifier Tree Search.}
To increase diversity, DVTS extends beam search by dividing the search process into $\frac{N}{M}$ subtrees, each of which is explored independently using beam search. As shown in~\citet{huggingface2024scaling}, DVTS outperforms beam search on easy and medium problems with a large computational budget $N$. A similar trend is observed in~\citet{AlphaMath}, where increasing the number of parallel subtrees proves to be more effective than increasing the beam width under the same budget.
% The detailed DVTS Algorithm can be found in Appendix~\ref{alg:dvts}.

\subsection{Compute-Optimal Test-Time Scaling}

To maximize the performance of TTS, \citet{snell2024scaling} proposes a test-time compute-optimal scaling strategy, which selects hyperparameters corresponding to a given test-time strategy to maximize performance benefits on a specific prompt. Given a prompt $x$, let $\operatorname{Target}(\theta, N, x)$ represent the output distribution over $x$ produced by the policy model with parameters $\theta$ and a compute budget of $N$.
\begin{equation}
\label{eq:compute_optimal}
    \theta^{*}_{x,y^*(x)}(N) = \underset{\theta}{\arg\max} \left( \mathbb{E}_{y \sim \operatorname{Target}(\theta, N, x)} \left[ \mathbbm{1}_{y = y^*(x)} \right] \right),
\end{equation}
where $y^*(x)$ denotes the ground-truth correct response for $x$, and $\theta^{*}_{x,y^*(x)}(N)$ represents the test-time compute-optimal scaling strategy for the problem $x$ with compute budget $N$.

\section{Rethinking Compute-Optimal Test-Time Scaling}\label{sec:rethink}

\subsection{Compute-Optimal Scaling Strategy Should be Reward-Aware}\label{sec:rethink_reward}

Compute-optimal TTS aims to allocate the optimal compute for each problem~\citep{snell2024scaling}. Previous works on TTS use a single \PRM as verifier~\citep{snell2024scaling, wu2024inference, huggingface2024scaling}. \citet{snell2024scaling} trains a \PRM on the responses of a policy model and uses it as the verifier to do TTS with the same policy model, while~\citet{wu2024inference, huggingface2024scaling} use a \PRM trained on a different policy model to do TTS. From the perspective of Reinforcement Learning~(RL), we obtain an \textit{on-policy} \PRM in the former case and an \textit{offline} \PRM in the latter case. The on-policy \PRM produces more accurate rewards for the responses of the policy model, while the offline \PRM often generates inaccurate rewards due to out-of-distribution (OOD) issues~\citep{snell2024scaling, ProcessBench}.

For practical applications of compute-optimal TTS, training a \PRM for each policy model to prevent OOD issues is computationally expensive. Therefore, we investigate the compute-optimal TTS strategy in a more general setting, where the \PRM might be trained on a different policy model than the one used for TTS. For search-based methods, \PRMs guide the selection at each response step, while for sampling-based methods, \PRMs evaluate the responses after generation. This indicates that (1) the reward influences response selection across all methods; (2) for search-based methods, the reward also influences the search process.

To analyze these points, we perform a preliminary case study using beam search with Llama-3.1-8B-Instruct as the policy model and RLHFlow-PRM-Mistral-8B and RLHFlow-PRM-Deepseek-8B as \PRMs. The results in Figure~\ref{fig:toycase} demonstrate that the reward significantly affects the generation process and outcomes. RLHFlow-PRM-Mistral-8B assigns high rewards to short responses, leading to incorrect answers, while searching with RLHFlow-Deepseek-PRM-8B produces correct answers but uses more tokens. In Section~\ref{sec:experiments_1}, we also empirically show that rewards have great influence on TTS performance and output tokens.

Based on these findings, we propose that \textit{rewards} should be integrated into the compute-optimal TTS strategy. Let us denote the reward function as $\mathcal{R}$. Our reward-aware compute-optimal TTS strategy is formulated as:
\begin{equation}
\label{eq:reward_aware_compute_optimal_tts}
    \theta^{*}_{x,y^*(x),\mathcal{R}}(N) = \underset{\theta}{\arg\max} \left( \mathbb{E}_{y \sim \operatorname{Target}(\theta, N, x, \mathcal{R})} \left[ \mathbbm{1}_{y = y^*(x)} \right] \right),
\end{equation}
where $\operatorname{Target}(\theta, N, x, \mathcal{R})$ represents the output distribution of the policy model $\theta$, adjusted by the reward function $\mathcal{R}$, under a compute budget $N$ and prompt $x$. For sampling-based scaling methods, $\operatorname{Target}(\theta, N, x, \mathcal{R}) = \operatorname{Target}(\theta, N, x)$. This reward-aware strategy ensures that compute-optimal scaling adapts to the policy model, prompt, and reward function, leading to a more general framework for practical TTS.

\subsection{Absolute Problem Difficulty Criterion is More Effective Than Quantiles}\label{sec:rethink_diff}

\begin{figure}[t]
\centering
\includegraphics[width=0.4\linewidth]{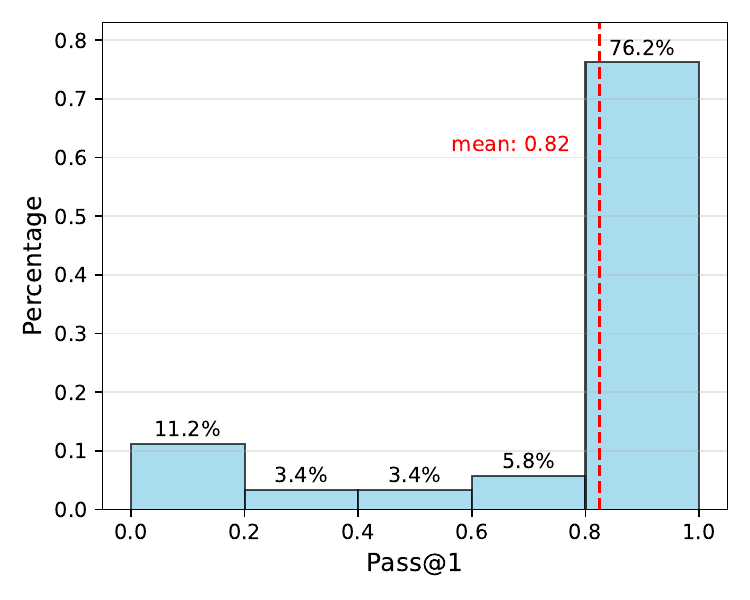}
\caption{Distribution of Pass@1 accuracy of Qwen2.5-72B-Instruct on MATH-500, divided into five bins.}
\label{fig:diff_bin}
\end{figure}

To consider the influence of problem difficulty on TTS, \citet{snell2024scaling} group problems into five difficulty levels based on Pass@1 accuracy quantiles. However, we find that using difficulty levels from MATH~\citep{MATH} or oracle labels based on Pass@1 accuracy quantiles~\citep{snell2024scaling} is not effective since different policy models have different reasoning capabilities. As shown in Figure~\ref{fig:diff_bin}, Qwen2.5-72B-Instruct achieves Pass@1 accuracy above $80\%$ on $76.2\%$ of MATH-500 problems. Therefore, we use absolute thresholds instead of quantiles to measure problem difficulty. Specifically, we define three difficulty levels based on Pass@1 accuracy: easy ($50\% \sim 100\%$), medium ($10\% \sim 50\%$), and hard ($0\% \sim 10\%$).

\section{How to Scale Test-Time Compute Optimally?}\label{sec:experiments_1}

In this section, we aim to answer the following questions:
\begin{itemize}
    \item \textbf{Q1}: How does TTS improve with different policy models and \PRMs?
    \item \textbf{Q2}: How does TTS improve for problems with different difficulty levels?
    \item \textbf{Q3}: Does \PRMs have bias towards specific response lengths or sensitivity to voting methods?
\end{itemize}

\subsection{Setup}

\paragraph{Datasets.}
We conduct experiments on competition-level mathematical datasets, including MATH-500~\citep{PRM800K} and AIME24~\citep{AIME24}. MATH-500 contains 500 representative problems from the test set of MATH~\citep{MATH}, and this subset is used following~\citet{snell2024scaling,huggingface2024scaling}. As recent LLMs show significant progress in mathematical reasoning~\citep{o1, DeepSeek-R1}, we include the more challenging AIME24 for experiments.

\paragraph{Policy Models.}
For test-time methods, we use policy models from Llama 3~\citep{Llama3} and Qwen2.5~\citep{Qwen2.5} families with different sizes. We use the \textit{Instruct} version for all policy models.

\paragraph{Process Reward Models.}
We consider the following open-source \PRMs for evaluation:
\begin{itemize}
    \item \textbf{Math-Shepherd}~\citep{Math-Shepherd}: \textbf{Math-Shepherd-PRM-7B} is trained on Mistral-7B~\citep{Mistral} using \PRM data generated from Mistral-7B fine-tuned on MetaMath~\citep{MetaMath}.
    \item \textbf{RLHFlow Series}~\citep{RLHFlow}: RLHFlow includes \textbf{RLHFlow-PRM-Mistral-8B} and \textbf{RLHFlow-PRM-Deepseek-8B}, which are trained on data from Mistral-7B fine-tuned on MetaMath~\citep{MetaMath} and deepseek-math-7b-instruct~\citep{DeepSeekMath}, respectively. The base model for both \PRMs is Llama-3.1-8B-Instruct~\citep{Llama3}.
    \item \textbf{Skywork Series}~\citep{Skywork-o1-open}: The Skywork series comprises \textbf{Skywork-PRM-1.5B} and \textbf{Skywork-PRM-7B}, trained on Qwen2.5-Math-1.5B-Instruct and Qwen2.5-Math-7B-Instruct~\citep{Qwen2.5-Math}, respectively. The training data is generated from Llama-2~\citep{Llama2} fine-tuned on a mathematical dataset and Qwen2-Math~\citep{Qwen2} series models.
    \item \textbf{Qwen2.5-Math Series}~\citep{PRMLessons}: We evaluate \textbf{Qwen2.5-Math-PRM-7B} and \textbf{Qwen2.5-Math-PRM-72B}, trained on Qwen2.5-Math-7B-Instruct and Qwen2.5-Math-72B-Instruct~\citep{Qwen2.5-Math}, respectively. The data for training is generated using Qwen2-Math~\citep{Qwen2} and Qwen2.5-Math series models~\citep{Qwen2.5-Math}. Among all the \PRMs listed, Qwen2.5-Math-PRM-72B is the strongest open-source \PRM for mathematical tasks, while Qwen2.5-Math-PRM-7B is the most capable \PRM among those with 7B/8B parameters, as demonstrated in~\citet{PRMLessons}.
\end{itemize}

% \section{Experimental Details}\label{app:exp_details}
% \paragraph{T}\label{app:exp_details}

\paragraph{Scoring and Voting Methods.}
Following~\citet{OpenR}, we consider three scoring methods: \textit{PRM-Min}, \textit{PRM-Last}, and \textit{PRM-Avg}, and three voting methods: \textit{Majority Vote}, \textit{PRM-Max}, and \textit{PRM-Vote}. To obtain the final answer, we first use the scoring methods to evaluate the answers. For a trajectory of length $H$, the scores for each trajectory with different scoring methods are calculated as follows: (1) \textit{PRM-Min} scores each trajectory by the minimum reward among all steps, i.e., $\operatorname{score} = \min_{\mathcal{R}}\{\mathcal{R}_t\}_{t=0}^{H}$. (2) \textit{PRM-Last} scores each trajectory by the reward of the last step, i.e., $\operatorname{score} = \mathcal{R}_H$. (3) \textit{PRM-Avg} scores each trajectory by the average reward among all steps, i.e., $\operatorname{score} = \frac{1}{H}\sum_{t=0}^{H}\mathcal{R}_t$. The voting methods then aggregate the scores to determine the final answer. \textit{Majority Vote} selects the answer with the majority of votes~\citep{Self-Consistency}, while \textit{PRM-Max} selects the answer with the highest score, and \textit{PRM-Vote} first accumulates the scores of all identical answers and then selects the answer with the highest score.

We use OpenR\footnote{\url{https://github.com/openreasoner/openr}}, which is an open-source LLM reasoning framework as our codebase. For compute budgets, we use $\{4, 16, 64, 256\}$ in most experiments. The division of steps follows the format \verb|\n\n| as in prior works~\citep{RLHFlow, PRMLessons}. For beam search and DVTS, the beam width is set to $4$. The temperature of CoT is $0.0$, while it is $0.7$ for other methods. For CoT and BoN, we restrict the maximum number of new tokens to $8192$. For search-based methods, the token limit is $2048$ for each step and $8192$ for the total response.

\subsection{How does TTS improve with different policy models and \PRMs ? (Q1)}

\begin{figure*}[!t]
\centering
\includegraphics[width=1.0\linewidth]{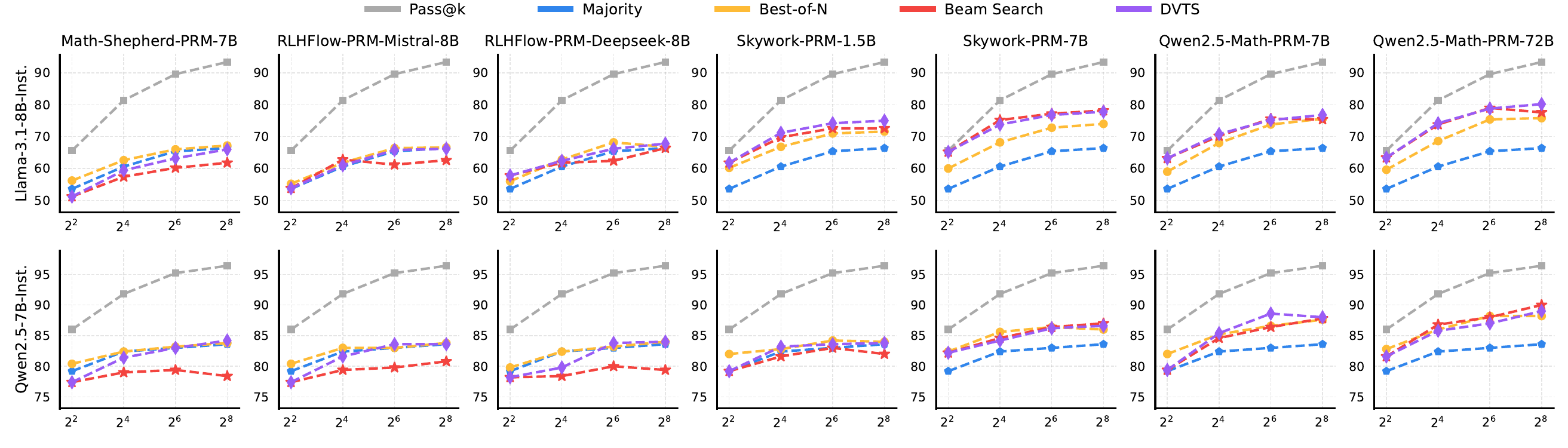}
\caption{Performance of Llama-3.1-8B-Instruct and Qwen2.5-7B-Instruct on MATH-500 with different \PRMs and TTS strategies.}
\label{fig:main_MATH}
\end{figure*}

\begin{figure*}[!t]
\centering
\includegraphics[width=1.0\linewidth]{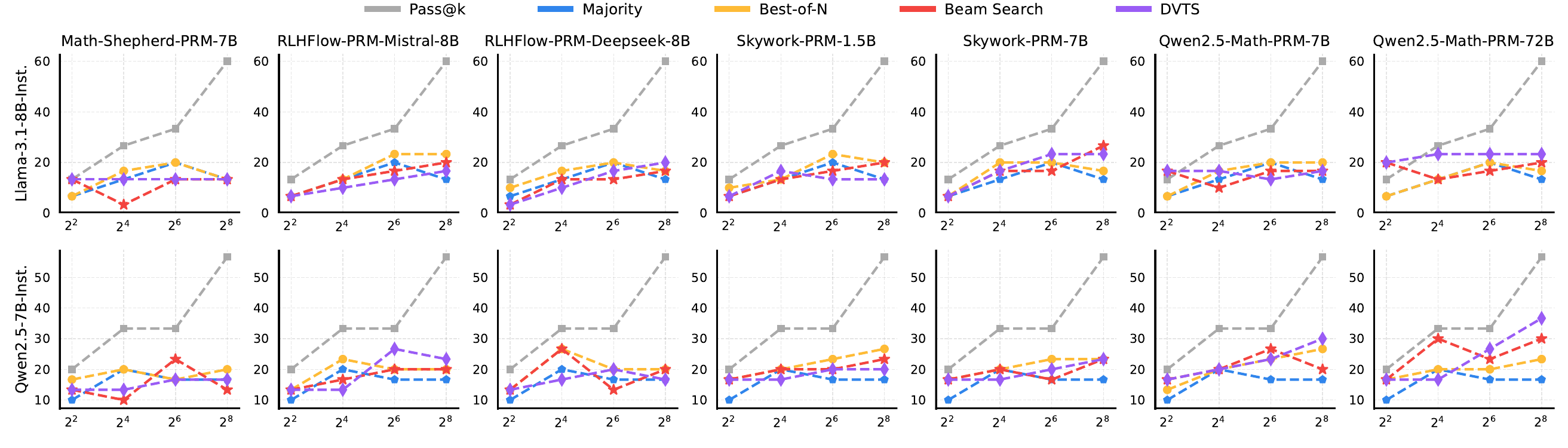}
\caption{Performance of Llama-3.1-8B-Instruct and Qwen2.5-7B-Instruct on AIME24 with different \PRMs and TTS strategies.}
\label{fig:main_AIME24}
\end{figure*}

\textbf{\PRMs are hard to generalize across policy models and tasks.}
As shown in Figure~\ref{fig:main_MATH}, for Llama-3.1-8B-Instruct, the performance of search-based methods with Skywork and Qwen2.5-Math \PRMs improves significantly with larger compute budgets, while the results of searching with Math-Shepherd and RLHFlow \PRMs remain relatively poor, even worse than majority voting.
For Qwen2.5-7B-Instruct, the performance of searching with Skywork-PRM-7B and Qwen2.5-Math \PRMs scales well with more budgets, while the performance of other \PRMs remains poor. In Figure~\ref{fig:main_AIME24}, although the Pass@k accuracy of both policy models improves a lot with larger compute budgets, the performance improvement of TTS remains moderate. These results demonstrate that the generalization of \PRMs is particularly challenging across different policy models and tasks, especially for more complex tasks.

\begin{figure}[!h]
\centering
\includegraphics[width=0.4\linewidth]{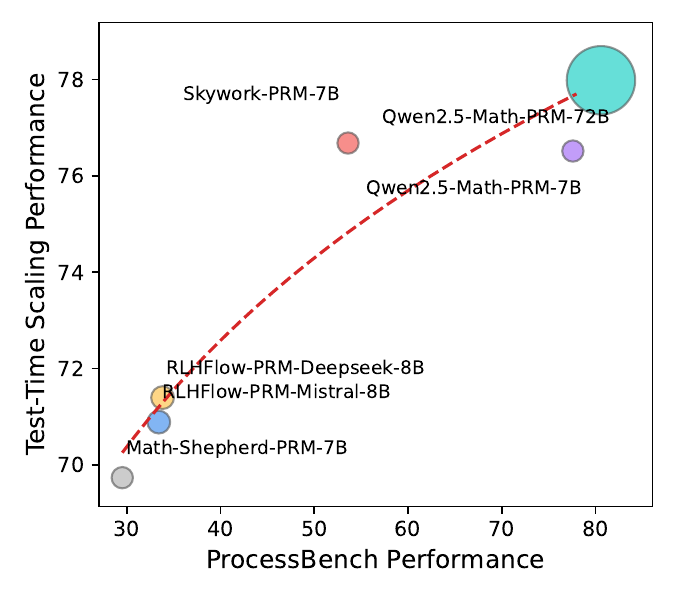}
\caption{The relationship between TTS performance and process supervision abilities of different \PRMs on MATH, where the size of each circle represents the number of parameters of the \PRM and the curve represents the fitted function.}
\label{fig:prm_processbench}
\end{figure}

\textbf{The optimal TTS method depends on the \PRM used.}
As shown in Figure~\ref{fig:main_MATH}, BoN outperforms other strategies most of the time when using Math-Shepherd and RLHFlow \PRMs, while search-based methods perform better with Skywork and Qwen2.5-Math \PRMs. This difference occurs because using a \PRM for OOD policy responses leads to sub-optimal answers, as \PRMs show limited generalization across policy models. Moreover, if we select \textit{each step} with OOD \PRMs, it is likely to obtain answers trapped in local optima and worsen the performance. This may also be related to the base model of the \PRM, since the \PRM trained with PRM800K~\citep{PRM800K} on Qwen2.5-Math-7B-Instruct generalizes better than \PRMs with Mistral and Llama as base models~\citep{PRMLessons}. Further analysis is provided in Section~\ref{app:exp_ablation} and Appendix~\ref{app:cases}. These results suggest that the choice of the optimal TTS strategy depends on the specific \PRMs used, emphasizing the importance of considering reward information in compute-optimal TTS. We also explore the relationship between TTS performance and the process supervision abilities of different \PRMs. As shown in Figure~\ref{fig:prm_processbench}, TTS performance is positively correlated with the process supervision abilities of \PRMs, and the fitted function is $Y = 7.66 \log(X) + 44.31$, where $Y$ represents TTS performance and $X$ represents the process supervision abilities of the \PRM~\citep{PRMLessons}.

\begin{figure*}[!h]
\centering
\includegraphics[width=1.0\linewidth]{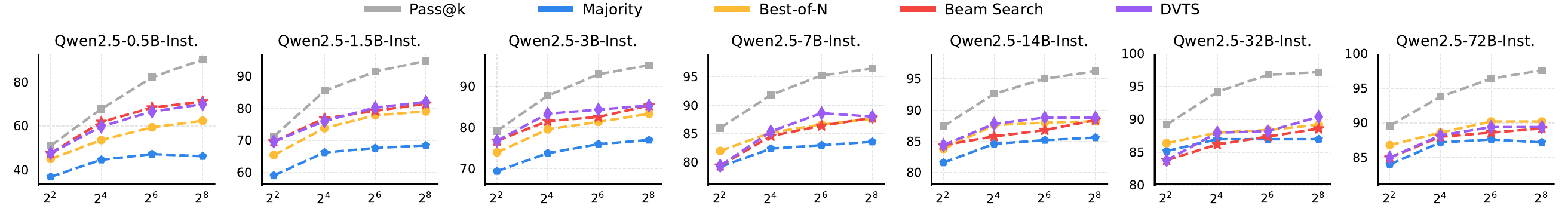}
\caption{TTS performance of policy models with parameters from 0.5B to 72B on MATH-500 with different scaling methods.}
\label{fig:policy_params}
\end{figure*}

\textbf{The optimal TTS method varies with policy models.}
To study the relationship between the parameters of the policy models and the optimal TTS methods, we conduct experiments with Qwen2.5 family LLMs~\citep{Qwen2.5}, including models with 0.5B, 1.5B, 3B, 7B, 14B, 32B, and 72B parameters. The results in Figure~\ref{fig:policy_params} show that the optimal TTS methods depend on the specific policy models. For small policy models, search-based methods outperform BoN, while for large policy models, BoN is more effective than search-based methods. This difference occurs because larger models have stronger reasoning capabilities and do not need a verifier to perform step-by-step selection. In contrast, smaller models rely on a verifier to select each step, ensuring the correctness of each intermediate step.
% More analysis can be found at Q5/Appendix (token, num steps).\revise{add more analysis}

\subsection{How does TTS improve for problems with different difficulty levels? (Q2)}

Following~\citet{snell2024scaling}, we conduct a comprehensive evaluation of tasks with varying difficulty levels. However, as explained in Section~\ref{sec:rethink_diff}, we observe that using the difficulty levels defined in MATH~\citep{MATH} or the oracle labels based on the quantile of Pass@1 accuracy~\citep{snell2024scaling} is not appropriate because different policy models exhibit different reasoning abilities. To address this, we categorize the difficulty levels into three groups based on the absolute value of Pass@1 accuracy: easy ($50\% \sim 100\%$), medium ($10\% \sim 50\%$), and hard ($0\% \sim 10\%$).

\begin{figure}[!h]
\centering
\includegraphics[width=0.55\linewidth]{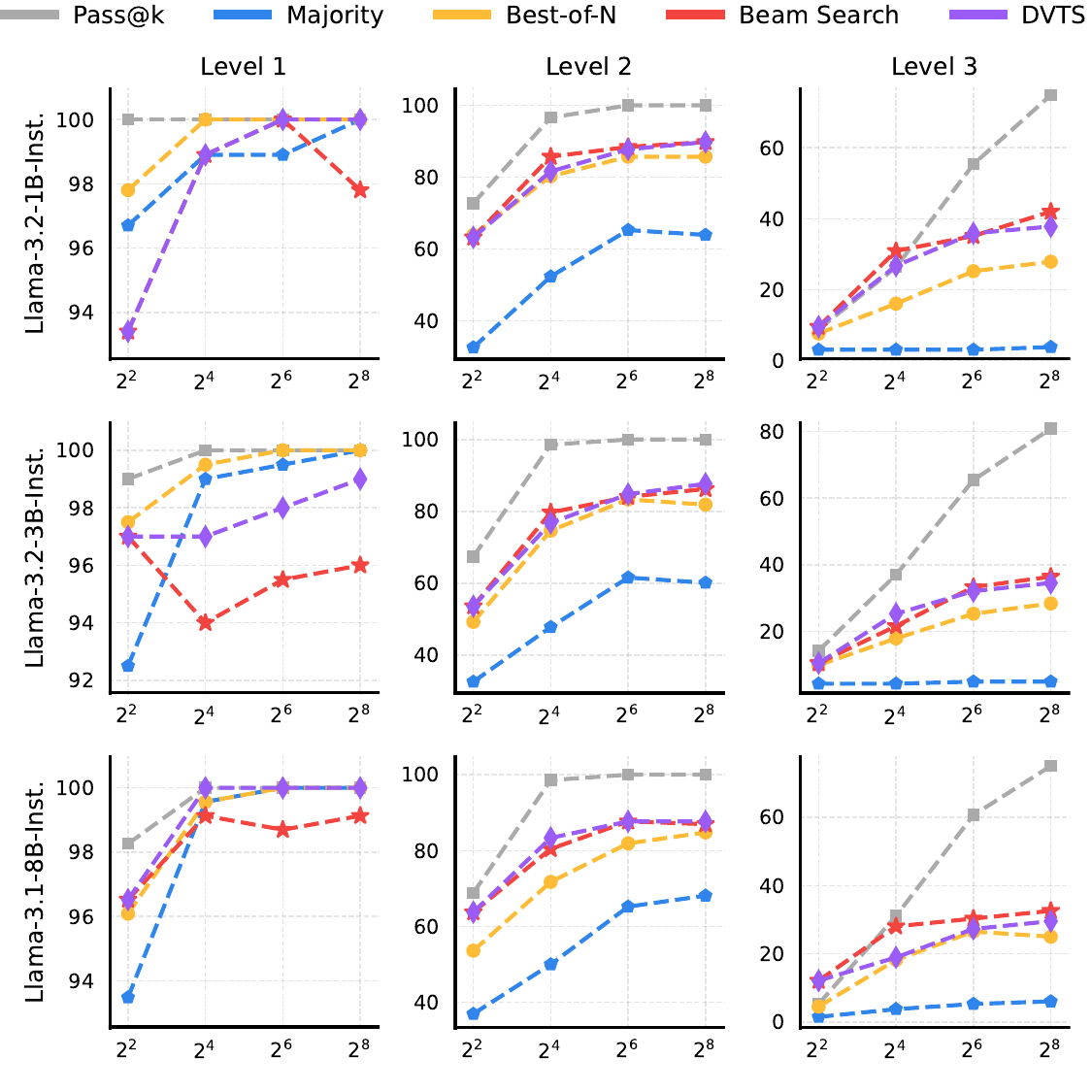}
\caption{TTS performance of three Llama policy models on MATH-500 with three difficulty levels.}
\label{fig:difficulty_main}
\end{figure}

\textbf{The optimal TTS methods vary with different difficulty levels.}
The results in Figure~\ref{fig:difficulty_main} and Figure~\ref{fig:difficulty_app} show that for small policy models (i.e., with fewer than 7B parameters), BoN is better for easy problems, while beam search works better for harder problems. For policy models with parameters between 7B and 32B, DVTS performs well for easy and medium problems, and beam search is preferable for hard problems. For policy models with 72B parameters, BoN is the best method for all difficulty levels.

\subsection{Does \PRMs have bias towards specific response lengths or sensitivity to voting methods? (Q3)}\label{app:exp_ablation}

\paragraph{\PRMs are biased towards the length of steps.}

\begin{table}[!h]
\centering
\caption{Statistics of training data of RLHFlow \PRMs.}
\resizebox{0.65\linewidth}{!}{
\begin{tabular}{lrr}
    \toprule
    & {\bf Mistral-PRM-Data} & {\bf Deepseek-PRM-Data} \\
    % & {\bf Mistral-Data} & {\bf Deepseek-Data} \\
    \midrule
    {Average Token per Response} & \mean{236.9} & \mean{333.1} \\
    {Average Token per Step}  & \mean{46.6} & \mean{58.4} \\
    \bottomrule
\end{tabular}
}
\label{tab:token}
\end{table}

Although we perform TTS under the same budget in pervious experiments, we find that the number of inference tokens with different \PRMs varies singificantly. For example, given the same budget and the same policy model, the number of inference tokens of scaling with RLHFlow-PRM-Deepseek-8B is consistently larger than that of RLHFlow-PRM-Mistral-8B, nearly $2\times$. The training data of RLHFlow series \PRMs are sampled from different LLMs, which may lead to the bias towards the length of the output. To verify this point, we analyze several properties of the training data of RLHFlow-PRM-Mistral-8B\footnote{\url{https://huggingface.co/datasets/RLHFlow/Mistral-PRM-Data}} and RLHFlow-PRM-Deepseek-8B\footnote{\url{https://huggingface.co/datasets/RLHFlow/Deepseek-PRM-Data}}. As shown in Table~\ref{tab:token}, both the average token per response and the average token per step of DeepSeek-PRM-Data are larger than those of Mistral-PRM-Data, indicating that the training data of RLHFlow-PRM-Deepseek-8B is longer than that of RLHFlow-PRM-Mistral-8B. This may lead to the bias towards the length of the output. We also find that the number of inference tokens of scaling with Qwen2.5-Math-7B is larger than that of Skywork-PRM-7B, but the performance is very near, which indicates that searching with Skywork-PRM-7B is more efficient than searching with Qwen2.5-Math-7B.

\paragraph{\PRMs are sensitive to voting methods.}

\begin{table}[!h]
\centering
\caption{Performance of TTS with different voting methods on MATH-500.}
\resizebox{0.55\linewidth}{!}{
\begin{tabular}{lrr}
    \toprule
    & {\bf Skywork-PRM-7B} & {\bf Qwen2.5-Math-PRM-7B} \\
    \midrule
    {\textit{Majority Vote}} & \mean{86.8} & \mean{87.6} \\
    {\textit{PRM-Min-Max}}   & \mean{83.0} & \mean{87.4} \\
    {\textit{PRM-Min-Vote}}  & \mean{86.6} & \mean{87.6} \\
    {\textit{PRM-Last-Max}}  & \mean{84.4} & \mean{87.6} \\
    {\textit{PRM-Last-Vote}} & \mean{87.0} & \mean{87.6} \\
    {\textit{PRM-Avg-Max}}   & \mean{85.8} & \mean{87.8} \\
    {\textit{PRM-Avg-Vote}}  & \mean{86.8} & \mean{87.6} \\
    \bottomrule
\end{tabular}
}
\label{tab:voting}
\end{table}

From the results in Table~\ref{tab:voting}, it is shown that Skywork-PRM-7B works better with \textit{PRM-Vote} than with \textit{PRM-Max}, while Qwen2.5-Math-PRM-7B is not very sensitive to voting methods. The main reason is that the training data of Qwen2.5-Math \PRMs are processed with LLM-as-a-judge~\citep{LLM-as-a-judge}, which removes the wrong intermediate steps labeled as positive steps in the training data and makes the outputted large reward values more likely to be correct. This shows that the training data of \PRMs is important for improving the ability to find errors in the search process.

\section{Results for Compute-Optimal Test-Time Scaling}\label{sec:experiments_2}

With the compute-optimal TTS strategy explored in Section~\ref{sec:experiments_1}, we conduct further experiments to explore the following questions:
\begin{itemize}
    \item \textbf{Q4}: Can smaller policy models outperform larger models with the compute-optimal TTS strategy?
    \item \textbf{Q5}: How does compute-optimal TTS improve compared with CoT and majority voting?
    \item \textbf{Q6}: Is TTS more effective than long-CoT-based methods?
\end{itemize}

\subsection{Can smaller policy models outperform larger models with the compute-optimal TTS strategy (Q4)}

Scaling test-time compute of small policy models is crucially important for improving the reasoning performance of LLMs. We are interested in whether smaller policy models can outperform larger ones, GPT-4o, even o1 and DeepSeek-R1, with the compute-optimal TTS strategy. First, we compare the performance of Llama-3.2-3B-Instruct (compute-optimal TTS) with that of Llama-3.1-405B-Instruct (CoT) on MATH-500 and AIME24. Also, we compare the performance of Qwen2.5-0.5B-Instruct, Qwen2.5-1.5B-Instruct, Llama-3.2-1B-Instruct, and Llama-3.2-3B-Instruct with GPT-4o on the above two tasks. As AIME24 is challenging for current LLMs, we also compare the performance of DeepSeek-R1-Distill-Qwen-1.5B and DeepSeek-R1-Distill-Qwen-7B with o1 on AIME24.

% Based on the compute-optimal TTS strategy, we compare the performance of small policy models with large policy models on MATH-500 and AIME24 under FLOPS matched settings.

\begin{table}[!t]
\centering
\caption{Comparison of small policy models (compute-optimal TTS) with frontier reasoning LLMs (CoT) on MATH-500 and AIME24.
% (Official results are marked with \official)
}
\resizebox{0.6\linewidth}{!}{
\begin{tabular}{lrrrrrr}
    \toprule
    {\bf Policy Model} & {\bf MATH-500} & {\bf AIME24} & {\bf Avg.} \\
    \midrule
    \multicolumn{4}{c}{\textit{Proprietary LLMs (CoT)}} \\
    \midrule
    {GPT-4o}     & \mean{74.6} & \mean{9.3} & \mean{42.0} \\
    {o1-preview} & \mean{85.5} & \mean{44.6} & \mean{65.1} \\
    {o1-mini}    & \mean{90.0} & \mean{63.6} & \mean{76.8} \\
    {o1}         & \mean{94.8} & \mean{79.2} & \mean{87.0} \\
    \midrule
    \multicolumn{4}{c}{\textit{Open-Source LLMs (CoT)}} \\
    \midrule
    {Llama-3.1-70B-Inst.}  & \mean{65.2} & \mean{16.7} & \mean{41.0} \\
    {Llama-3.1-405B-Inst.} & \mean{71.4} & \mean{23.3} & \mean{47.4} \\
    {QwQ-32B-Preview}         & \mean{90.6} & \mean{50.0} & \mean{70.3} \\
    {DeepSeek-R1}             & \mean{97.3} & \mean{79.8} & \mean{88.6} \\
    \midrule
    \multicolumn{4}{c}{\textit{Open-Source LLMs (TTS)}} \\
    \midrule
    {Llama-3.2-1B-Inst.} & \mean{66.2} & \mean{16.7} & \mean{41.5} \\
    {Llama-3.2-1B-Inst. ($N=512$)} & \mean{72.2} & \mean{10.0} & \mean{41.1} \\
    {Llama-3.2-3B-Inst.} & \mean{75.6} & \mean{30.0} & \mean{52.8} \\
    {Qwen2.5-0.5B-Inst.} & \mean{76.4} & \mean{10.0} & \mean{43.2} \\
    {Qwen2.5-1.5B-Inst.} & \mean{81.8} & \mean{20.0} & \mean{50.9} \\
    {DeepSeek-R1-Distill-Qwen-1.5B} & \mean{91.6} & \mean{63.3} & \mean{77.5} \\
    {DeepSeek-R1-Distill-Qwen-7B}   & \mean{95.2} & \mean{83.3} & \mean{89.3} \\
    \bottomrule
\end{tabular}
}
\label{tab:small_vs_large}
\end{table}

From the results in Table~\ref{tab:small_vs_large}, we have the following observations: (1) Llama-3.2-3B-Instruct with the compute-optimal TTS strategy outperforms Llama-3.1-405B-Instruct on MATH-500 and AIME24, meaning that \textbf{smaller models can outperform 135$\times$ larger models using the compute-optimal TTS strategy}. Compared with previous works on TTS~\citep{snell2024scaling, huggingface2024scaling}, we improve the result by \textbf{487.0\%} ($23\times \rightarrow 135\times$).
(2) If we further increase the compute budget to $N=512$, Llama-3.2-1B-Instruct with the compute-optimal TTS strategy beats Llama-3.1-405B-Instruct on MATH-500, but underperforms Llama-3.1-405B-Instruct on AIME24.\footnote{Since some outputs of Llama-3.2-1B-Instruct do not contain \texttt{\textbackslash boxed}, which is used for answer extraction, we use Qwen2.5-32B-Instruct to extract the answers of Llama-3.2-1B-Instruct.}
(3) Qwen2.5-0.5B-Instruct and Llama-3.2-3B-Instruct with the compute-optimal TTS strategy outperforms GPT-4o, indicating that \textbf{small models can exceed GPT-level performance with the compute-optimal TTS strategy}.
(4) DeepSeek-R1-Distill-Qwen-1.5B with the compute-optimal TTS strategy outperforms o1-preview and o1-mini on MATH-500 and AIME24. We also show that DeepSeek-R1-Distill-Qwen-7B with the compute-optimal TTS strategy outperforms o1 and DeepSeek-R1 on MATH-500 and AIME24. These results demonstrate \textbf{small reasoning-enhanced models can outperform frontier reasoning LLMs with the compute-optimal TTS strategy}.

\paragraph{FLOPS Comparison.}
To answer the question of whether compute-optimal TTS is more effective than increasing the model size, we compare the FLOPS of evaluated models in Table~\ref{tab:small_vs_large_FLOPS} following~\citet{snell2024scaling}, where the computed FLOPS is corresponded to the results in Table~\ref{tab:small_vs_large}. From the results, we can see that \textbf{small policy models even surpass large ones  with less inference FLOPS} and reduce the total FLOPS by $100\times \sim 1000\times$.
% \liu{lack conclusion}

\begin{table}[!t]
\centering
\caption{FLOPS comparison between smaller policy models (compute-optimal TTS) and larger ones (CoT).
}
\resizebox{0.75\linewidth}{!}{
\begin{tabular}{lrrrrrr}
    \toprule
    {\bf Policy Model} & {\bf Pre-training FLOPS} & {\bf Inference FLOPS} & {\bf Total FLOPS.} \\
    \midrule
    {Llama-3.2-3B-Inst.}   & \mean{$1.62 \times 10^{23}$} & \mean{$3.07 \times 10^{17}$} & \mean{$1.62 \times 10^{23}$} \\
    {Llama-3.1-405B-Inst.} & \mean{$3.65 \times 10^{25}$} & \mean{$4.25 \times 10^{17}$} & \mean{$3.65 \times 10^{25}$} \\
    \midrule
    {DeepSeek-R1-Distill-7B}   & \mean{$7.56 \times 10^{23}$} & \mean{$8.15 \times 10^{17}$} & \mean{$7.56 \times 10^{23}$} \\
    {DeepSeek-R1} & \mean{$5.96 \times 10^{25}$} & \mean{$4.03 \times 10^{18}$} & \mean{$5.96 \times 10^{25}$} \\
    \bottomrule
\end{tabular}
}
\label{tab:small_vs_large_FLOPS}
\end{table}

\subsection{How does compute-optimal TTS improve compared with CoT and majority voting? (Q5)}

Based on the findings of compute-optimal TTS with different policy models, \PRMs, and difficulty levels, we summarize the results of compute-optimal TTS for each policy model on MATH-500 in Table~\ref{tab:cot_vs_majority_vs_co}. We find that compute-optimal TTS can be $256 \times$ more efficient than majority voting and improve reasoning performance by $154.6\%$ over CoT. These results demonstrate that compute-optimal TTS significantly enhances the reasoning capabilities of LLMs. However, as the number of parameters in the policy model increases, the improvement of TTS gradually decreases. This suggests that the effectiveness of TTS is directly related to the reasoning ability of the policy model. Specifically, for models with weak reasoning abilities, scaling test-time compute leads to a substantial improvement, whereas for models with strong reasoning abilities, the gain is limited.

\begin{table}[!t]
\centering
\caption{Comparison of compute-optimal TTS, CoT, and majority voting with different policy models on MATH-500.
}
\resizebox{0.85\linewidth}{!}{
\begin{tabular}{lrrrrr}
    \toprule
    {\bf Policy Model} & {\bf CoT} & {\bf Major.} & {\bf Compute-Optimal TTS} & {\bf Performance Gain} & {\bf Efficiency Gain} \\
    \midrule
    {Llama-3.2-1B-Inst.} & \mean{26.0} & \mean{39.0} & \mean{66.2} & \mean{154.6\%} & \mean{>256.0\tms} \\
    {Llama-3.2-3B-Inst.} & \mean{41.4} & \mean{58.4} & \mean{78.2} & \mean{88.9\%} & \mean{14.1\tms} \\
    {Llama-3.1-8B-Inst.} & \mean{49.8} & \mean{66.4} & \mean{80.6} & \mean{61.8\%} & \mean{43.9\tms} \\
    \midrule
    {Qwen2.5-0.5B-Inst.} & \mean{31.6} & \mean{47.2} & \mean{76.4} & \mean{141.8\%} & \mean{>64.0\tms} \\
    {Qwen2.5-1.5B-Inst.} & \mean{54.4} & \mean{68.4} & \mean{85.6} & \mean{57.4\%} & \mean{>256.0\tms} \\
    {Qwen2.5-3B-Inst.} & \mean{64.0} & \mean{77.0} & \mean{87.6} & \mean{36.9\%} & \mean{58.4\tms} \\
    {Qwen2.5-7B-Inst.} & \mean{76.8} & \mean{83.6} & \mean{91.0} & \mean{18.5\%} & \mean{35.9\tms} \\
    {Qwen2.5-14B-Inst.} & \mean{80.2} & \mean{85.6} & \mean{91.0} & \mean{13.5\%} & \mean{51.4\tms} \\
    {Qwen2.5-32B-Inst.} & \mean{82.4} & \mean{87.0} & \mean{90.6} & \mean{10.0\%} & \mean{0.8\tms} \\
    {Qwen2.5-72B-Inst.} & \mean{83.8} & \mean{87.2} & \mean{91.8} & \mean{9.5\%} & \mean{12.9\tms} \\
    \bottomrule
\end{tabular}
}
\label{tab:cot_vs_majority_vs_co}
\end{table}

\subsection{Is TTS more effective than long-CoT-based methods? (Q6)}

Recently, long-CoT-based methods have shown substantial progress in mathematical reasoning~\citep{rStar-Math, PRIME, SimpleRL, DeepSeek-R1}. We compare the performance of TTS with these approaches.

\paragraph{Setup.}
We evaluate the following methods: (1) \textbf{rStar-Math}~\citep{rStar-Math}: This method first generates reasoning data via MCTS, followed by online policy and preference model learning. (2) \textbf{Eurus-2}~\citep{PRIME}: This method enhances the reasoning abilities of LLMs through implicit process rewards and online RL. (3) \textbf{SimpleRL}~\citep{SimpleRL}: This method replicates self-reflection with only 8K training data. (4) \textbf{Satori}~\citep{Satori}: This method first learn the format and then improves the reasoning abilities via RL. (5) \textbf{DeepSeek-R1-Distill-Qwen-7B}~\citep{DeepSeek-R1}: This method distills 800K high-quality reasoning samples from DeepSeek-R1 with 671B parameters into a 7B LLM.

\paragraph{Results.}
As shown in Table~\ref{tab:long-cot}, we find that TTS with Qwen2.5-7B-Instruct outperforms rStar-Math, Eurus-2, SimpleRL, and Satori on both MATH-500 and AIME24. However, while the performance of TTS on MATH-500 is close to that of DeepSeek-R1-Distill-Qwen-7B, it shows a significant drop on AIME24. These results indicate that TTS is more effective than methods applying direct RL or SFT on the data generated via MCTS but is less effective than distilling from strong reasoning models. Also, TTS is more effective on simpler tasks than on more complex tasks.

\begin{table}[!t]
\centering
\caption{Comparison of compute-optimal TTS with long-CoT methods on MATH-500 and AIME24.
}
\resizebox{0.65\linewidth}{!}{
\begin{tabular}{lrrrrrr}
    \toprule
    {\bf Policy Model} & {\bf MATH-500} & {\bf AIME24} & {\bf Avg.} \\
    \midrule
    \multicolumn{4}{c}{\textit{Open-Source LLMs (CoT)}} \\
    \midrule
    {Qwen2.5-7B-Inst.}         & \mean{76.8} & \mean{13.3} & \mean{45.1} \\
    {Qwen2.5-Math-7B-Inst.}    & \mean{79.8} & \mean{13.3} & \mean{46.6} \\
    \midrule
    \multicolumn{4}{c}{\textit{Long-CoT Methods (CoT)}} \\
    \midrule
    {rStar-Math-7B}               & \mean{78.4} & \mean{26.7} & \mean{52.6} \\
    {Eurus-2-7B-PRIME}            & \mean{79.2} & \mean{26.7} & \mean{53.0} \\
    {Qwen2.5-7B-SimpleRL-Zero}    & \mean{77.2} & \mean{33.3} & \mean{55.3} \\
    {Qwen2.5-7B-SimpleRL}         & \mean{82.4} & \mean{26.7} & \mean{54.6} \\
    {Satori-Qwen-7B}              & \mean{83.6} & \mean{23.3} & \mean{53.5} \\
    {DeepSeek-R1-Distill-Qwen-7B} & \mean{92.4} & \mean{63.3} & \mean{77.9} \\
    \midrule
    \multicolumn{4}{c}{\textit{Open-Source LLMs (TTS)}} \\
    \midrule
    {Qwen2.5-7B-Inst. w/ 7B PRM (Ours)} & \mean{88.0} & \mean{33.3} & \mean{60.5} \\
    {Qwen2.5-7B-Inst. w/ 72B PRM (Ours)} & \mean{91.0} & \mean{36.7} & \mean{63.9} \\
    \bottomrule
\end{tabular}
}
\label{tab:long-cot}
\end{table}

\section{Related Work}

\paragraph{LLM Test-Time Scaling.}

Scaling LLM test-time compute is an effective way to improve the performance~\citep{o1}. Previous works explore majority voting~\citep{Self-Consistency}, search-based methods~\citep{ToT, xie2024self, ARGS, wan2024alphazero}, and refinement~\citep{RISE} to improve the performance. For verification-guided test-time compute, \citet{brown2024large} explores inference compute with repeated sampling and domain verifiers, while \citet{MindStar, wu2024inference, snell2024scaling} further explore search-based methods with process reward guidance and \citet{xiyao2024scaling} extends this setting to VLMs. To eliminate the need for external reward models and the generation of extensive samples, \citet{manvi2024adaptive} proposes a self-evaluation method for adaptive and efficient test-time compute. A recent work~\citep{huggingface2024scaling} explores TTS via search methods with diversity. However, these works lack a evaluation with either strong verifiers or policies with different sizes / capabilities.
In this paper, we aim to provide a more systematically evaluation with up-to-date policies and verifiers, more challenging tasks, and provide some principles for practical TTS.

\paragraph{Improving Mathematical Reasoning Abilities of LLMs.}

Prior methods for improving mathematical reasoning abilities can be divided into training-time methods and test-time methods.
For training-time methods, previous works explore large-scale mathematical corpus pre-training~\citep{GPT-4, Llemma, DeepSeekMath} and supervised fine-tuning~\citep{Wizardmath, MetaMath, ToRA, MathScale, DART-Math, Skywork-Math} to improve mathematical capabilities.
Another line of works explore self-training and self-improvement strategies~\citep{STaR, ReST, ReFT, V-STaR, Quiet-STaR, ReST-MCTS*, setlur2024rl, kumar2024training, PRIME}, which improve the reasoning abilities by fine-tuning on self-generated solutions.
% \paragraph{o1-related.}
Recently, many works improve the mathematical reasoning abilities with long CoT~\citep{o1-Journey1, o1-Journey2, k0-math, DeepSeek-R1, QwQ, Skywork-o1, Marco-o1}, as OpenAI o1~\citep{o1} shows significantly powerful reasoning capabilities with long thinking.

For test-time methods, prompt-based approaches have been extensively studied to enhance reasoning without altering the model parameters. Techniques such as Chain-of-Thought (CoT)~\citep{CoT} and its variants~\citep{ToT, CoMAT} guide the model to decompose problems into manageable sub-steps, thereby improving accuracy and coherence in mathematical reasoning. Beyond prompting strategies, self-refinement techniques~\citep{Self-Refine} allow models to review and correct their outputs, while external tool integration~\citep{PAL, PoT} leverages program interpreter or symbolic manipulators to perform precise calculations and validations.
Self-verification approaches~\citep{Self-Verification} enable models to assess the correctness of their own reasoning processes, further increasing robustness.
% Moreover, ensemble methods and multi-agent collaborations~\citep{ensemble_methods, multi_model} aggregate diverse reasoning paths to reach more accurate conclusions.
These test-time strategies complement training-time enhancements, collectively contributing to significant improvements in LLMs' mathematical reasoning capabilities. Our work mainly enhances the reasoning performance via scaling test-time compute via PRM-guided search methods.

\paragraph{Process Reward Models.}

Previous works show that \PRMs are more effective than \ORMs~\citep{uesato2022solving, PRM800K}. However, collecting high-quality \PRMs data, such as PRM800K~\citep{PRM800K}, is often costly. 
The researchers explores automatic \PRM data collection via direct Monte Carlo estimation~\citep{Math-Shepherd}, detecting relative scores of \ORMs~\citep{AutoPSV}, and efficient MCTS with binary search~\citep{OmegaPRM}.
Recently, more advanced \PRMs are explored from advantage modeling~\citep{PAV}, $Q$-value rankings~\citep{PQM}, implicit rewards~\citep{Implicit-PRM}, and entropy regularization~\citep{ER-PRM} perspectives. Additionally, more open-source \PRMs are released~\citep{RLHFlow, Skywork-o1, ER-PRM, PQM, Implicit-PRM, PRMLessons}, showing strong performance on mathematical tasks. With the rapid development of \PRMs, ProcessBench~\citep{ProcessBench} and PRMBench~\citep{PRMBench} are proposed to provide comprehensive evaluation of \PRMs.~\citet{PRMLessons} provides guidelines for practical development of \PRMs and releases the most capable \PRMs for mathematical tasks up-to-date.

% \citet{ma2023let} PRM search

\section{Conclusion \& Discussion}
In this paper, we present a thorough empirical analysis of compute-optimal test-time scaling from the perspectives of different policy models, \PRMs, and more challenging evaluation tasks. Our findings demonstrate the dependency of compute-optimal TTS strategies on policy models, \PRMs, and problem difficulty, validating that smaller language models can perform better than larger models when applying compute-optimal TTS. Our results show that a 1B model can achieve better performance than a 405B model through TTS. Additionally, we demonstrate that a 7B PRM can achieve strong TTS results by supervising a more capable 72B policy model, which suggests the importance of investigating a true ``weak-to-strong'' approach instead of the current ``strong-to-weak'' supervision for policy optimization. To achieve this goal, we need to develop more efficient supervision methods, as both PRM-based and RL-based approaches have limitations due to their dependence on high-quality supervision. Future work should focus on developing more adaptable and universal supervision mechanisms to boost the performance of small language models on complex tasks and provide new approaches for developing efficient reasoning strategies.

\paragraph{Limitations.}
Although we provide a comprehensive evaluation of TTS on mathematical tasks, there are still some limitations and future directions to explore: (1) Extending TTS to more tasks such as coding and chemistry tasks. (2) Exploring more effective methods for compute-optimal TTS.

\clearpage
\bibliography{arxiv}

\clearpage
\appendix

\section{Prompt Template for Test-Time Scaling}

The system prompt for Llama 3 series models~\citep{Llama3} and Qwen2.5 series models~\citep{Qwen2.5} are listed in Table~\ref{tab:llama_prompt} and Table~\ref{tab:qwen_prompt}, respectively. Following~\citet{huggingface2024scaling}, we use the system prompt of the official evaluation\footnote{\url{https://huggingface.co/datasets/meta-llama/Llama-3.2-1B-Instruct-evals}} for Llama 3 to prevent performance drop.

\begin{table*}[!ht]
\centering
\caption{System prompt for Llama 3 series models.}
\begin{tcolorbox}[
    colframe=black, 
    colback=white, 
    boxrule=1pt, 
    arc=0pt, 
    fontupper=\small\ttfamily,
    boxsep=3pt,
    left=5pt,
    right=0pt,
    parbox=false
]
\begin{lstlisting}[
    breaklines, 
    basicstyle=\small\ttfamily,
    frame=none,
    backgroundcolor=\color{white},
    escapeinside={(*}{*)},
    captionpos=b,
    columns=fullflexible,
    keepspaces=true,
    xleftmargin=0pt,
    xrightmargin=0pt
]
Solve the following math problem efficiently and clearly:

- For simple problems (2 steps or fewer):
Provide a concise solution with minimal explanation.

- For complex problems (3 steps or more):
Use this step-by-step format:

## Step 1: [Concise description]
[Brief explanation and calculations]

## Step 2: [Concise description]
[Brief explanation and calculations]

...

Regardless of the approach, always conclude with:

Therefore, the final answer is: $\boxed{answer}$. I hope it is correct.

Where [answer] is just the final number or expression that solves the problem.
\end{lstlisting}
\end{tcolorbox}
\label{tab:llama_prompt}
\end{table*}

\begin{table*}[!ht]
\centering
\caption{System prompt for Qwen2.5 series models.}
\begin{tcolorbox}[
    colframe=black, 
    colback=white, 
    boxrule=1pt, 
    arc=0pt, 
    fontupper=\small\ttfamily,
    boxsep=3pt,
    left=5pt,
    right=0pt,
    parbox=false
]
\begin{lstlisting}[
    breaklines, 
    basicstyle=\small\ttfamily,
    frame=none,
    backgroundcolor=\color{white},
    escapeinside={(*}{*)},
    captionpos=b,
    columns=fullflexible,
    keepspaces=true,
    xleftmargin=0pt,
    xrightmargin=0pt
]
Please reason step by step, and put your final answer within \boxed{}.
\end{lstlisting}
\end{tcolorbox}
\label{tab:qwen_prompt}
\end{table*}

\section{Full Results of Test-Time Scaling with Different Policy Models, PRMs, and Scaling Methods}

\begin{figure*}[!ht]
\centering
% \vspace*{-0.5em}
\includegraphics[width=1.0\linewidth]{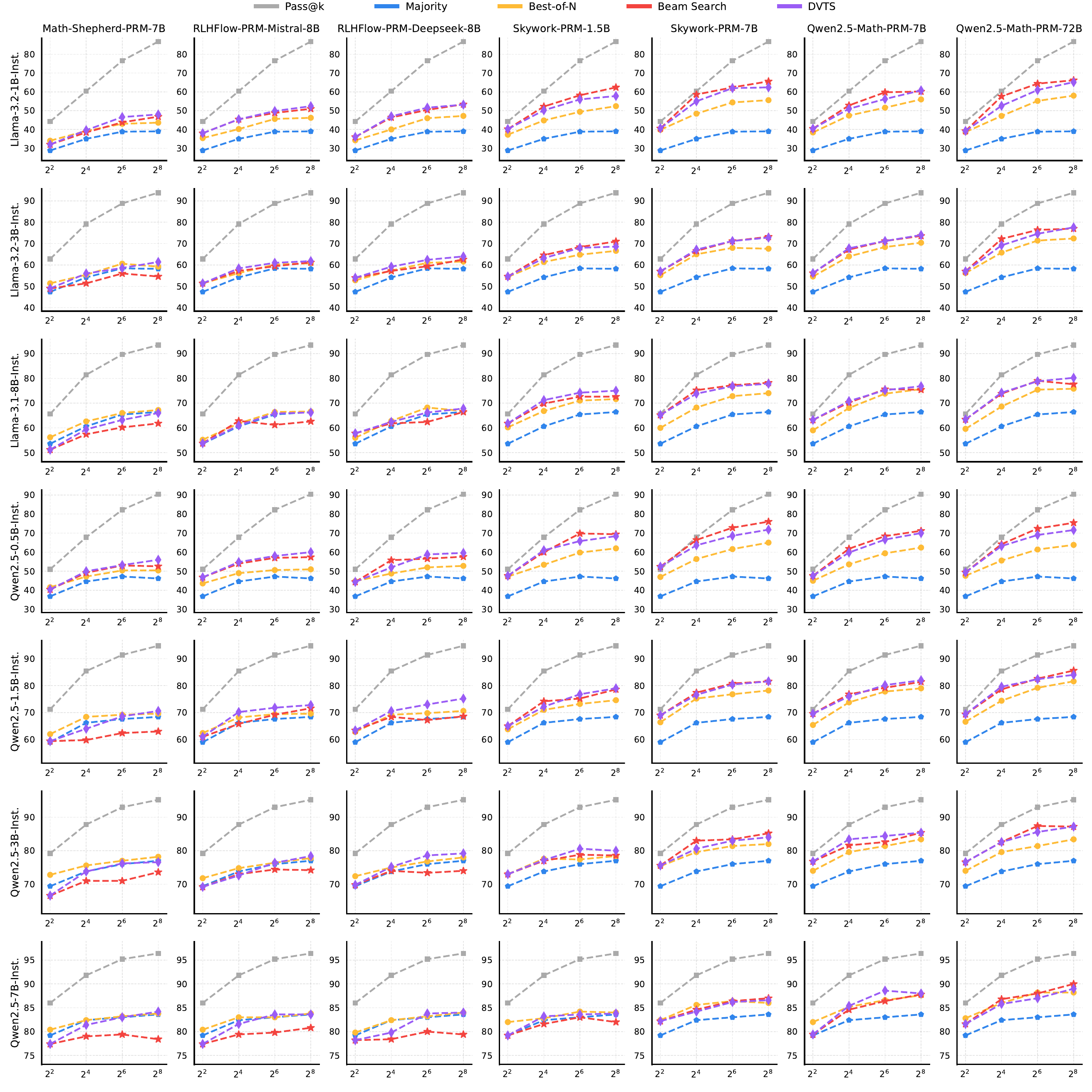}
\caption{TTS performance of different policy models on MATH-500 with different \PRMs and scaling strategies.}
\label{fig:app_MATH}
\end{figure*}

\begin{figure*}[!ht]
\centering
% \vspace*{-0.5em}
\includegraphics[width=1.0\linewidth]{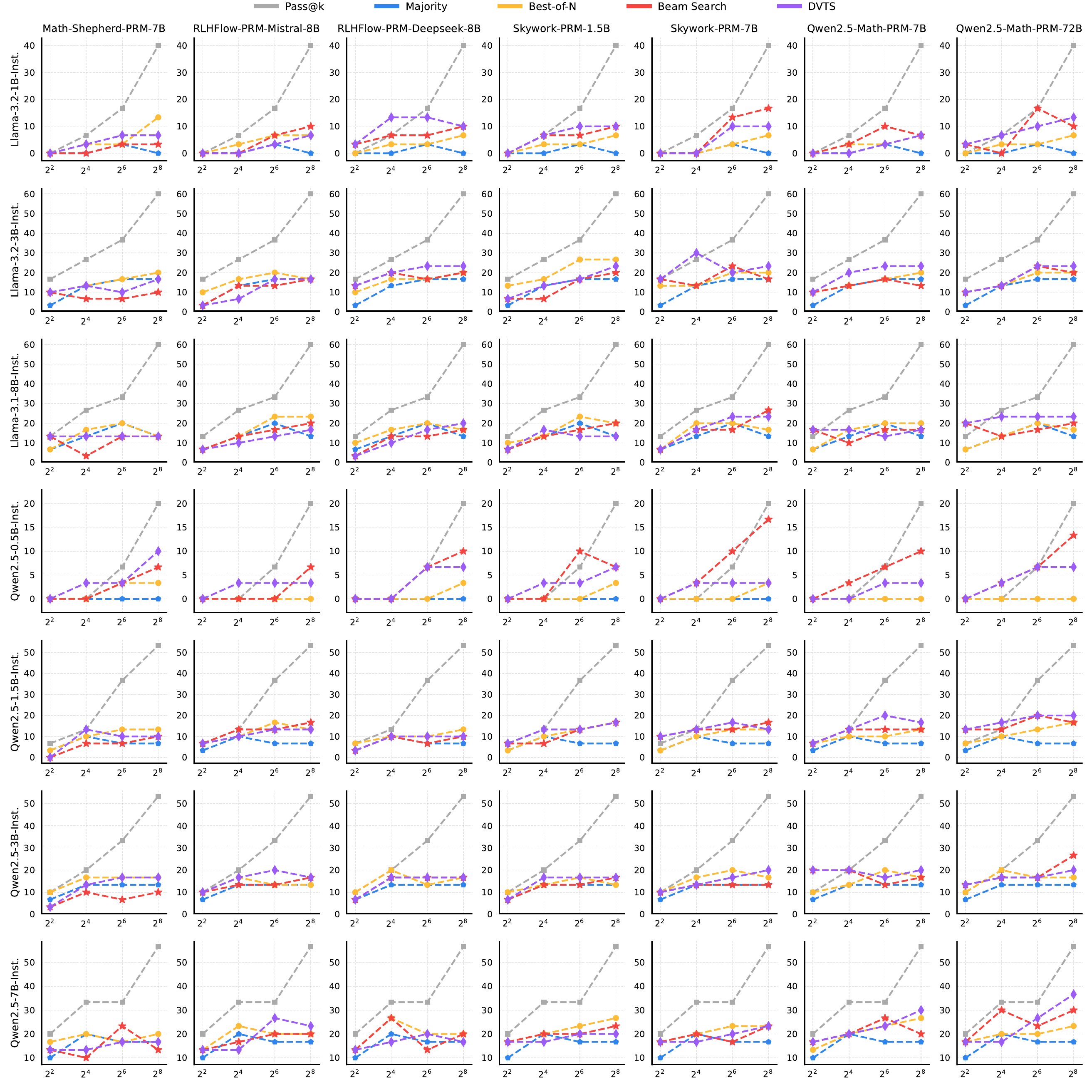}
\caption{TTS performance of different policy models on AIME24 with different \PRMs and scaling strategies.}
\label{fig:app_AIME24}
\end{figure*}

The full results of TTS with different policy models, PRMs, and scaling methods are shown in Figure~\ref{fig:app_MATH} and Figure~\ref{fig:app_AIME24}.

\begin{figure*}[!ht]
\centering
\vspace{-0.5em}
\includegraphics[width=0.4\linewidth]{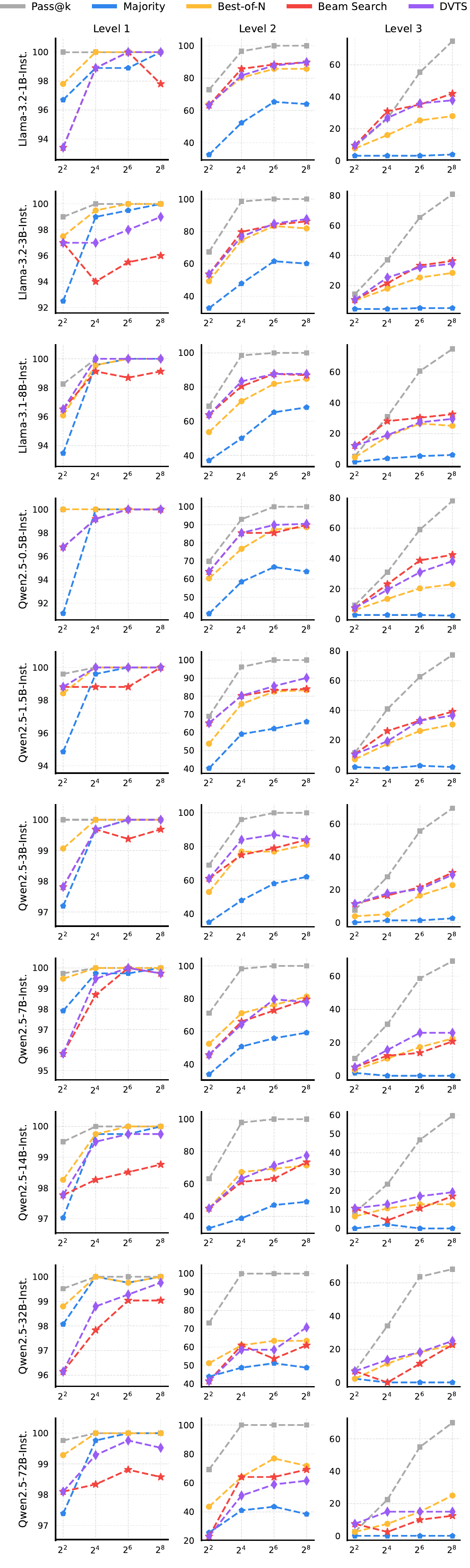}
\vspace{-0.5em}
\caption{TTS performance of three Llama policy models on MATH-500 with different difficulty levels.}
\label{fig:difficulty_app}
\end{figure*}

\clearpage

\section{Cases}\label{app:cases}

\begin{figure}[!ht]
\centering
\includegraphics[width=0.8\linewidth]{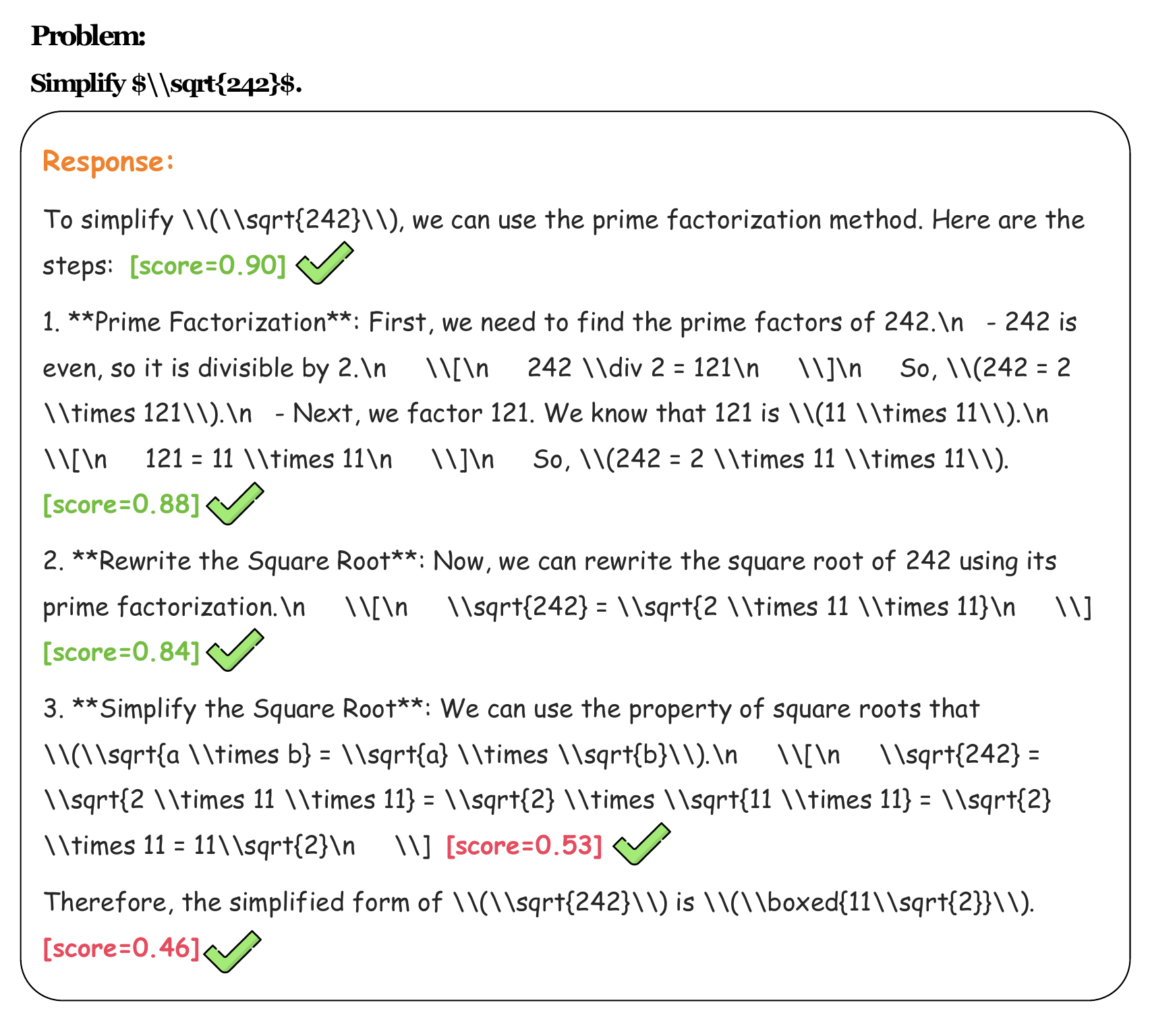}
\caption{TTS case of Over-Criticism.}
\label{fig:case1}
\end{figure}

\begin{figure}[!ht]
\centering
\includegraphics[width=0.8\linewidth]{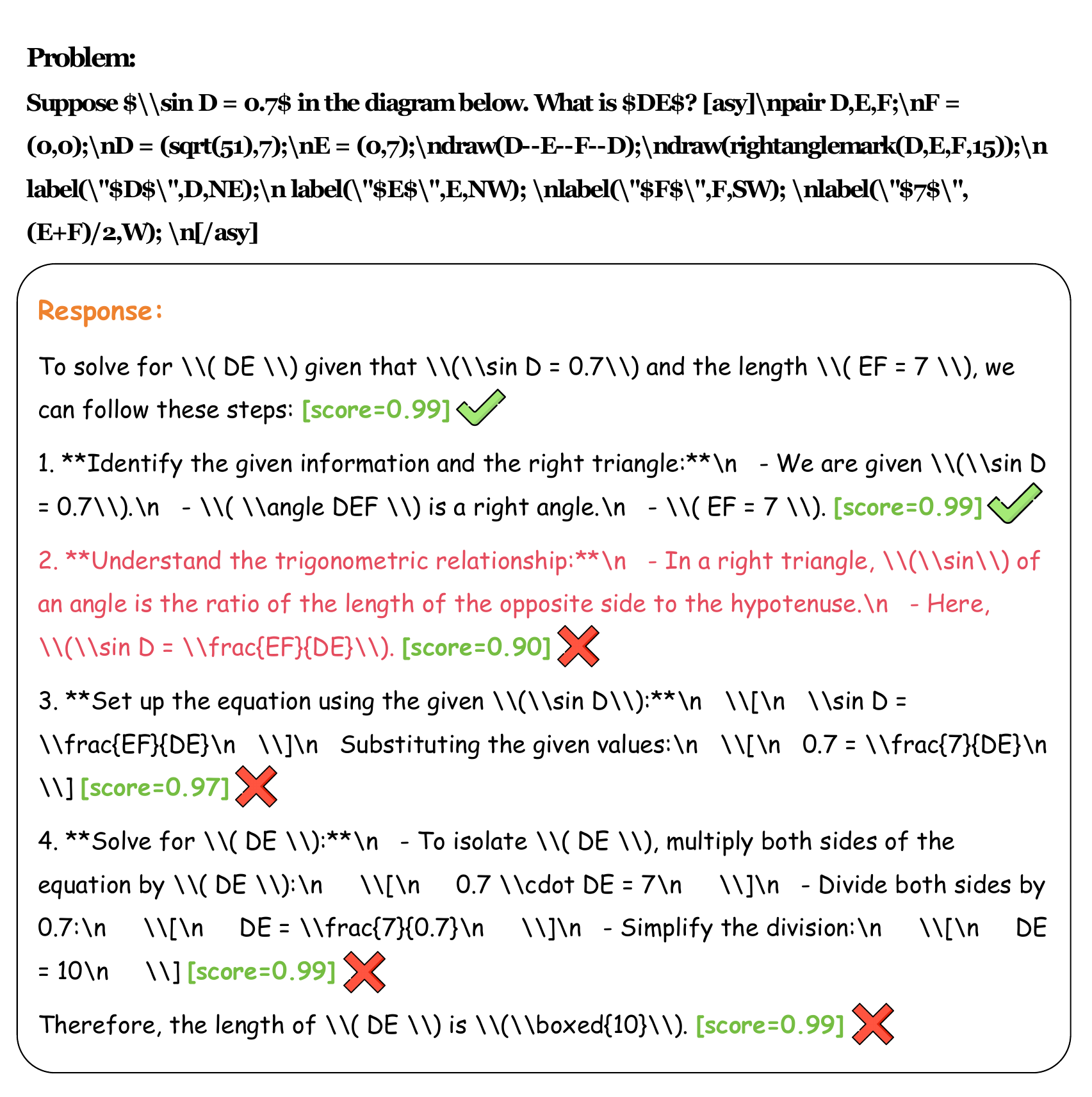}
\caption{TTS case of Error Neglect.}
\label{fig:case2}
\end{figure}

\begin{figure}[!ht]
\centering
\includegraphics[width=0.8\linewidth]{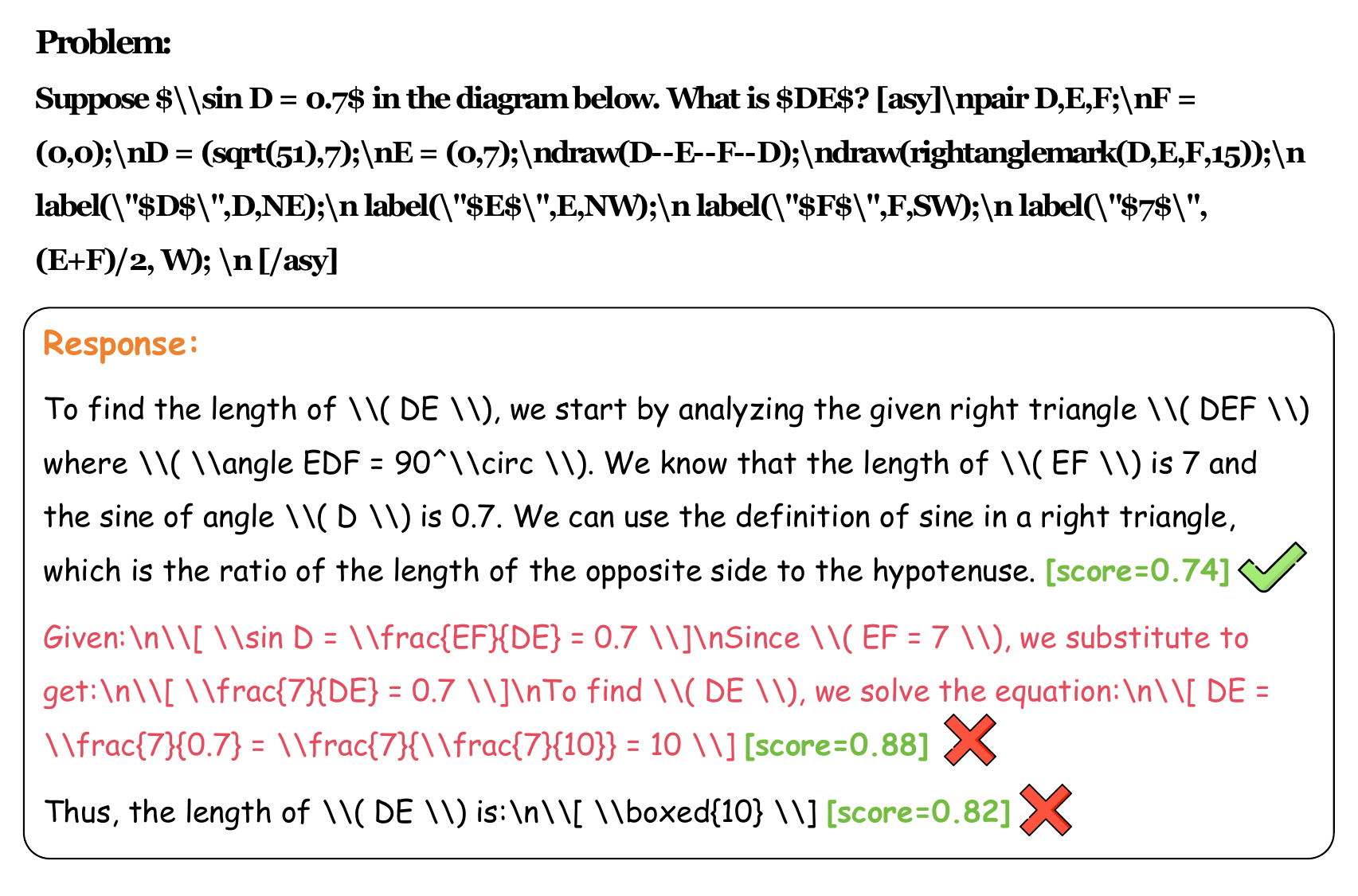}
\caption{TTS case of Error Neglect.}
\label{fig:case3}
\end{figure}

\begin{figure}[!ht]
\centering
\includegraphics[width=0.8\linewidth]{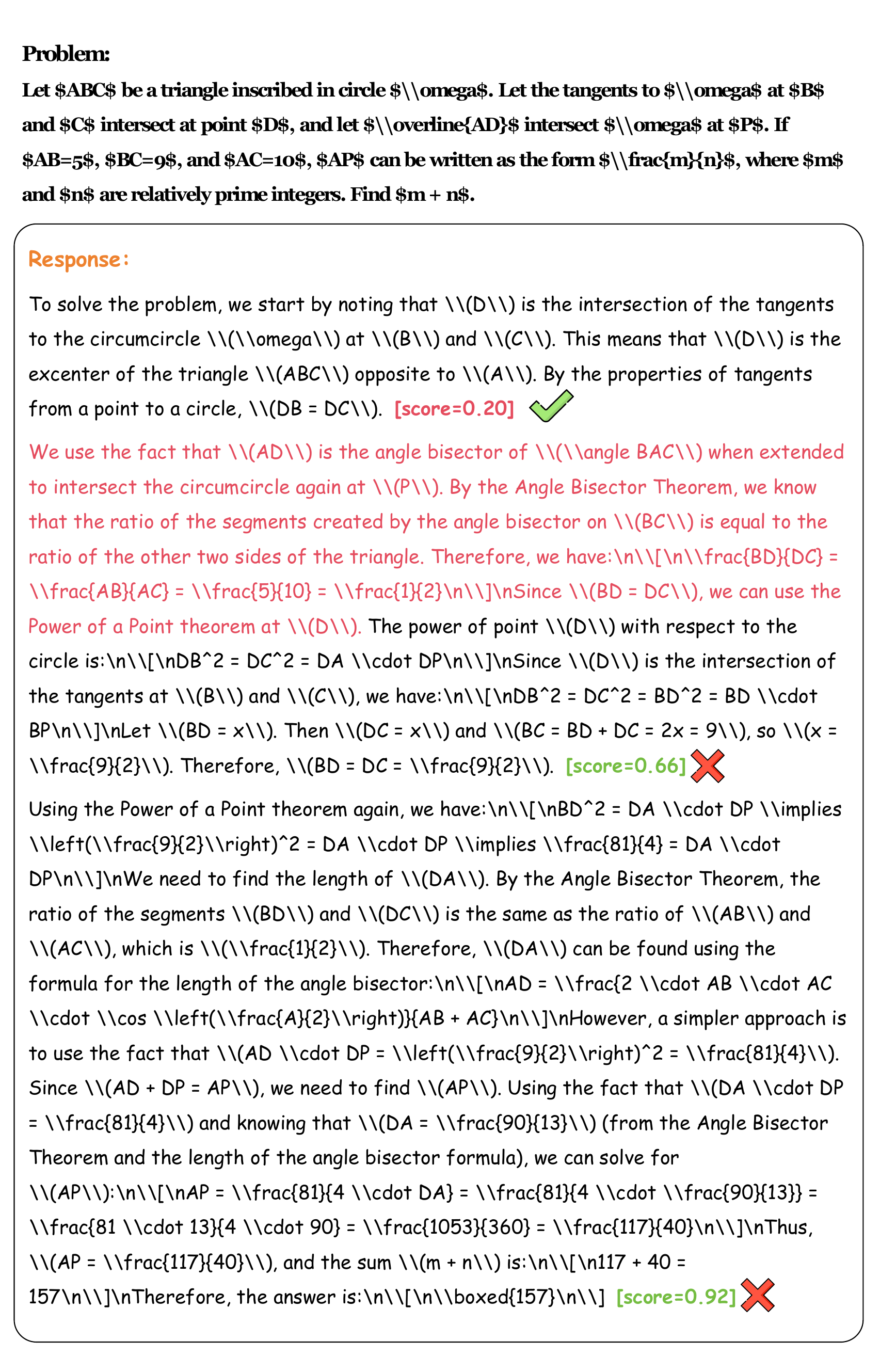}
\caption{TTS case of Error Localization Bias.}
\label{fig:case4}
\end{figure}

\begin{figure}[!ht]
\centering
\includegraphics[width=0.8\linewidth]{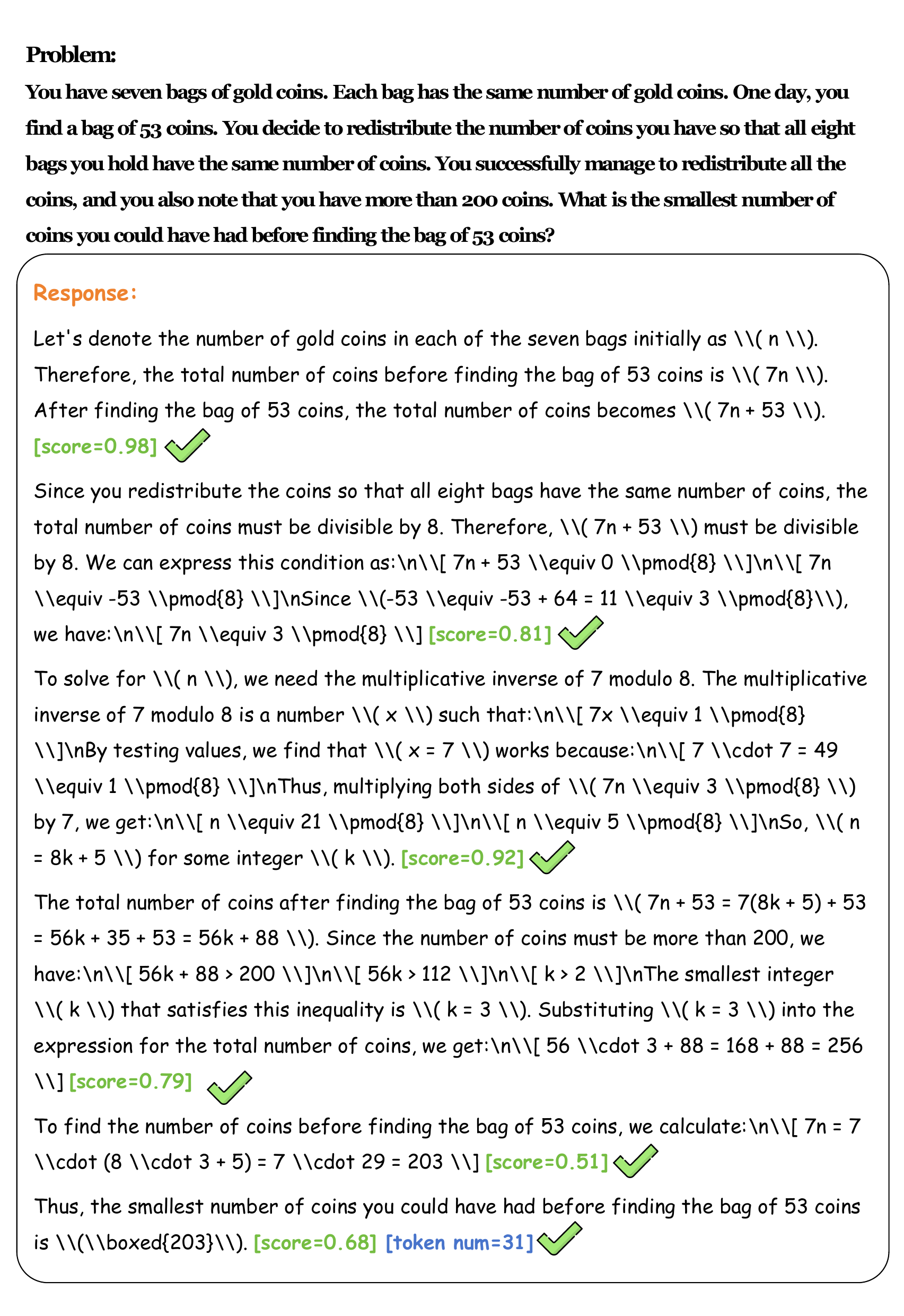}
\caption{TTS case of Scoring Bias.}
\label{fig:case5}
\end{figure}

\begin{figure}[!ht]
\centering
\includegraphics[width=0.8\linewidth]{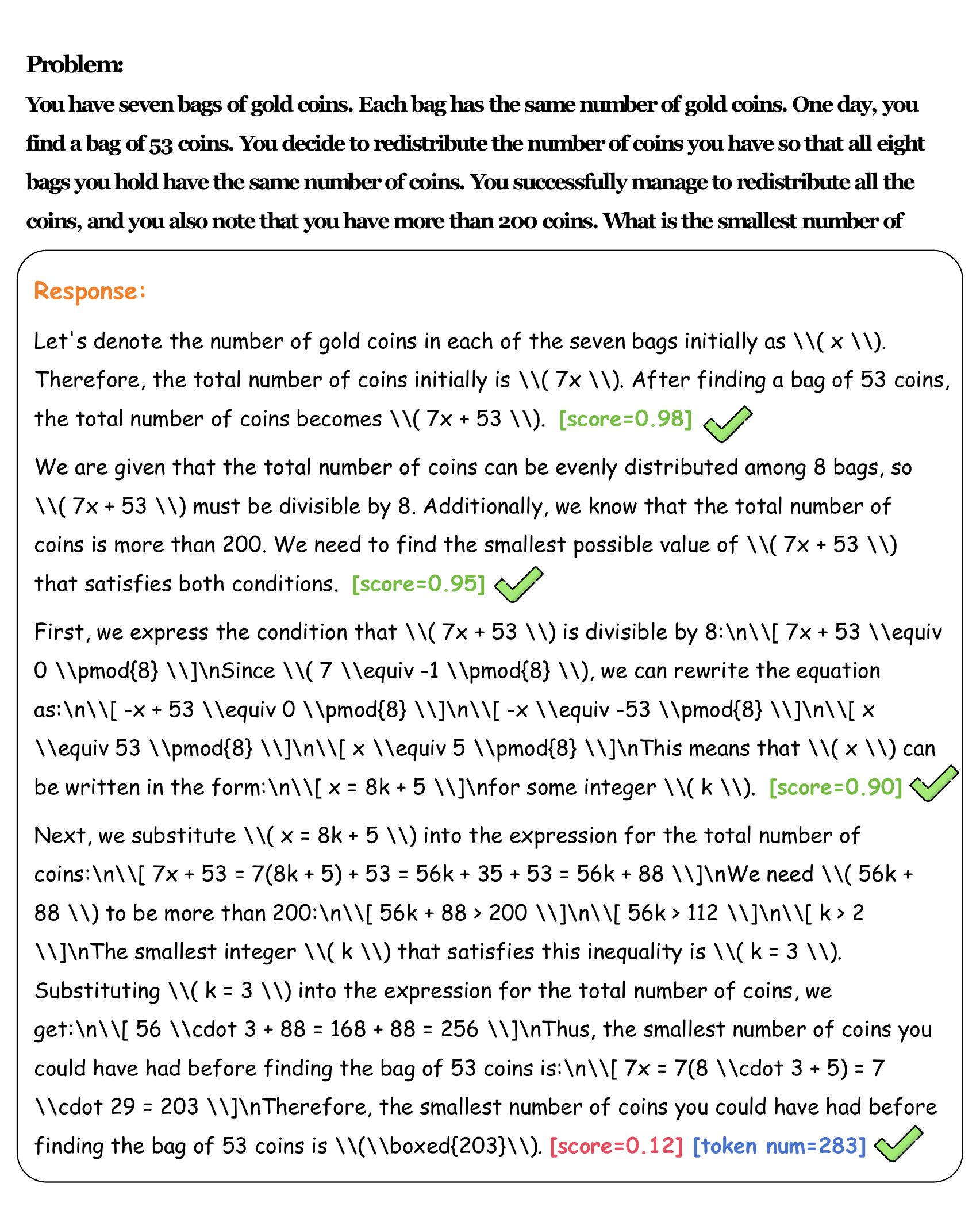}
\caption{TTS case of Scoring Bias.}
\label{fig:case6}
\end{figure}

In this section, we provide cases for TTS and summarize several problems for \PRMs. By analyzing the output of TTS, we identify several issues with \PRMs. Specifically, we observe four major categories:
(1) \textbf{Over-Criticism}: As shown in Figure~\ref{fig:case1}, the \PRM assigns low scores even to mathematically correct steps, resulting in false negatives.
(2) \textbf{Error Neglect}: As shown in Figure~\ref{fig:case2} and Figure~\ref{fig:case3}, the \PRM sometimes assigns relatively high scores to steps with clear mathematical errors, failing to detect these errors during the reasoning process.
(3) \textbf{Error Localization Bias}: As shown in Figure~\ref{fig:case4}, the \PRM assigns lower scores to certain intermediate steps that are not where the critical errors actually occur. This indicates a misalignment between the scoring signal and the actual error locations.
(4) \textbf{Scoring Bias}: As shown in Figure~\ref{fig:case5} and Figure~\ref{fig:case6}, certain training biases, such as sensitivity to the token length of intermediate steps, result in large discrepancies in scores for equally correct reasoning steps.

Notably, these issues persist across both OOD datasets (e.g., the AIME24 dataset, which was not used during \PRM training) and in-distribution data (e.g., the MATH dataset used to train the model). These problems distort the reasoning search process, degrade overall performance, and reduce the reliability of \PRM-assisted reasoning. Addressing these biases in future model architectures and training procedures is necessary to improve the robustness and interpretability of \PRMs.

\begin{figure}[!ht]
\centering
\includegraphics[width=0.8\linewidth]{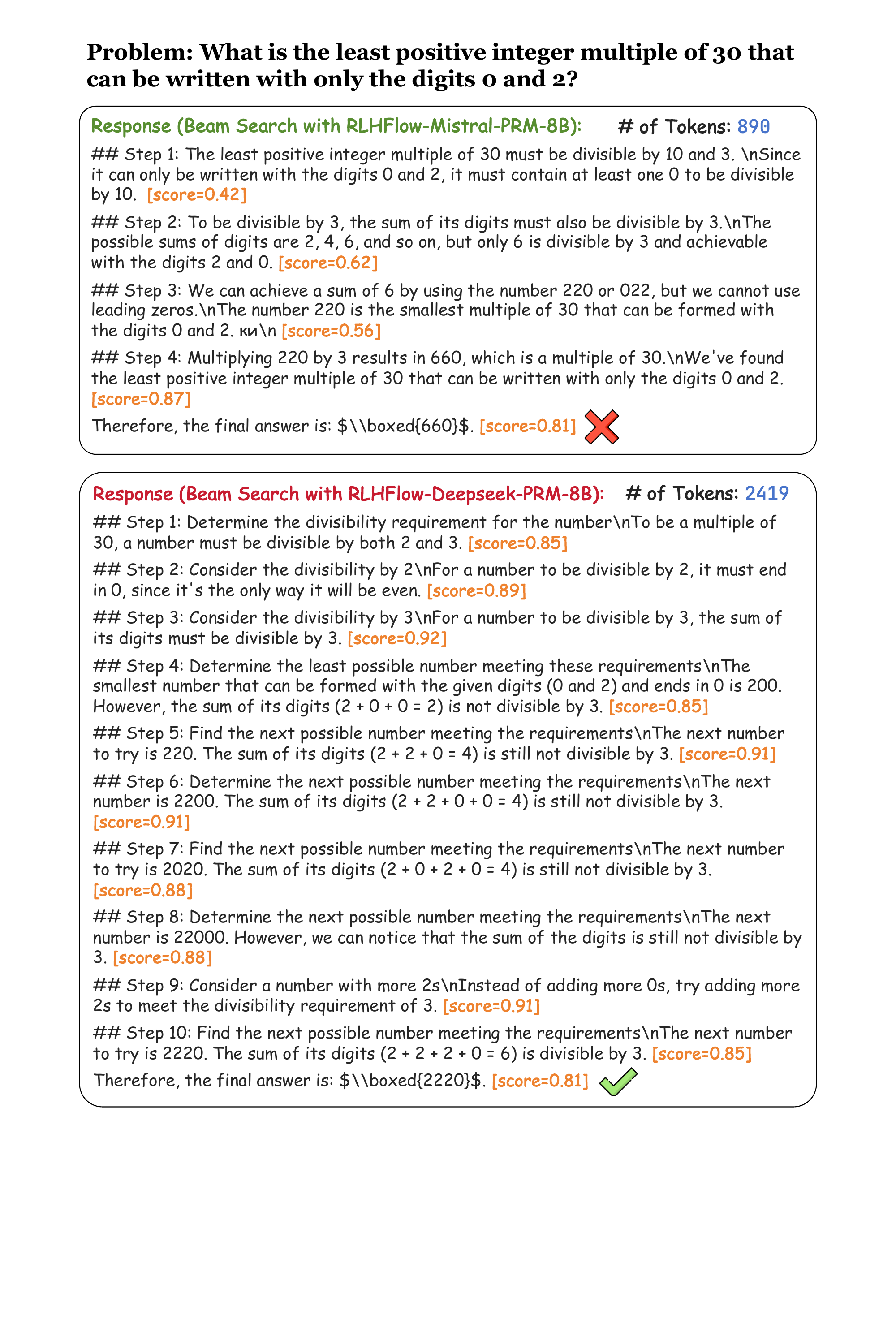}
\caption{Toy case of beam search with RLHFlow-Mistral-PRM-8B and RLHFlow-Deepseek-PRM-8B.}
\label{fig:toycase}
\end{figure}

\end{document}